\documentclass[lettersize,journal]{IEEEtran}
\usepackage{amsmath,amsfonts}
\usepackage{algorithm}
\usepackage{algpseudocode}
\usepackage{array}
\usepackage[caption=false,font=normalsize,labelfont=sf,textfont=sf]{subfig}
\usepackage{textcomp}
\usepackage{stfloats}
\usepackage{url}
\usepackage{verbatim}
\usepackage{graphicx}
\usepackage{tcolorbox}
\tcbuselibrary{most}
\usepackage{multicol}
\usepackage{cite}
\usepackage{multirow}
\usepackage{pifont}
\usepackage{booktabs}
\usepackage{comment}
\usepackage{algpseudocode}

\usepackage{hyperref}

\usepackage{orcidlink}

\newcommand{\sysname}{{DreamScene}}
\newtcolorbox{mybox}[2][]{colbacktitle=red!5!white, colback= blue!5!white,coltitle=red!70!black, title={#2},fonttitle=\bfseries,#1}
\begin{document}
\title{DreamScene: 3D Gaussian-based End-to-end Text-to-3D Scene Generation} 
\author{
\thanks{This work was supported by Anhui Province Science and Technology Innovation Breakthrough Plan (202423l10050033) and the National Key Research and Development Program of China (2022YFB3105405, 2021YFC3300502). Corresponding author: Yong Liao.}
Haoran Li, Yuli Tian, Kun Lan, Yong Liao*~\IEEEmembership{Member,~IEEE}, Lin Wang~\IEEEmembership{Member,~IEEE}, Pan Hui~\IEEEmembership{Fellow,~IEEE}, Peng Yuan Zhou~\IEEEmembership{Member,~IEEE}
\thanks{Haoran Li, Yuli Tian, Kun Lan and Yong Liao are with University of Science and Technology of China, Hefei, China (e-mail: lhr123@mail.ustc.edu.cn; yltian@mail.ustc.edu.cn; lankun@mail.ustc.edu.cn; yliao@ustc.edu.cn).}
\thanks{Lin Wang is with the School of Electrical and Electronic Engineering, Nanyang Technological University, Singapore (email:eee-addison.wang@ntu.edu.sg).}
\thanks{Pan Hui is with the Computational Media and Arts thrust, Hong Kong University of Science and Technology (Guangzhou), China, and Department of Computer Science, University of Helsinki, Finland (email:panhui@ust.hk).}
\thanks{Peng Yuan Zhou is with the Department of Electrical and Computer Engineering, Aarhus University, Denmark (email: pengyuan.zhou@ece.au.dk).}

}

\markboth{IEEE TRANSACTIONS ON PATTERN ANALYSIS AND MACHINE INTELLIGENCE}%
{Shell \MakeLowercase{\textit{et al.}}: A Sample Article Using IEEEtran.cls for IEEE Journals}


\maketitle




\begin{abstract}
\textcolor{black}{Generating 3D scenes from natural language holds great promise for applications in gaming, film, and design. However, existing methods struggle with automation, 3D consistency, and fine-grained control. We present \sysname, an end-to-end framework for high-quality and editable 3D scene generation from text or dialogue. \sysname\ begins with a scene planning module, where a GPT-4 agent infers object semantics and spatial constraints to construct a hybrid graph. A graph-based placement algorithm then produces a structured, collision-free layout. Based on this layout, Formation Pattern Sampling (FPS) generates object geometry using multi-timestep sampling and reconstructive optimization, enabling fast and realistic synthesis. To ensure global consistent, \sysname\ employs a progressive camera sampling strategy tailored to both indoor and outdoor settings. Finally, the system supports fine-grained scene editing, including object movement, appearance changes, and 4D dynamic motion. Experiments demonstrate that \sysname\ surpasses prior methods in quality, consistency, and flexibility, offering a practical solution for open-domain 3D content creation. Code and demos are available at \url{https://jahnsonblack.github.io/DreamScene-Full/}.}

\end{abstract}




\section{introduction}
\label{sec:intro}
\IEEEPARstart{T}{he} progress made in text-to-3D scene generation 
signifies a significant step forward in the field of 3D content creation~\cite{poole2022dreamfusion,lin2023magic3d,chen2023fantasia3d,liu2023zero,metzer2023latent,huang2023dreamtime,yu2023text,liang2023luciddreamer,tang2023dreamgaussian,li2023sweetdreamer,nichol2022point,jun2023shap}. It has extended its reach from generating simple objects to building intricate, detailed scenes straight from the textual descriptions. This advancement not only lightens the burden on 3D modelers but also stimulates expansion in industries like gaming, film, and architecture.

\par
\begin{figure*}[t!]
\centering
\includegraphics[width=\textwidth]{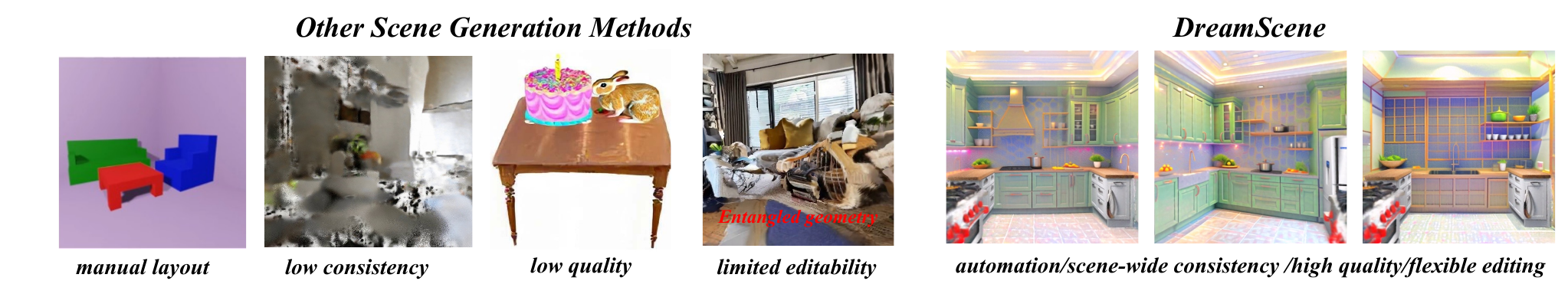}
\vspace{-0pt}
\caption{\textcolor{black}{\sysname\ exhibits significant advantages compared with current state-of-the-art text-to-3D scene generation methods.
Text2Room~\cite{hollein2023text2room} and Set-the-Scene~\cite{cohen2023set}  require complex user-specified object placement.
Text2Room, Text2NeRF~\cite{zhang2024text2nerf} and many inpainting-based methods suffer from low scene consistency, producing incoherent geometry across camera poses. GALA3D~\cite{zhou2024gala3d}, CG3D~\cite{vilesov2023cg3d} generate scenes with low visual quality and do not generate 3D environments.
Moreover, most existing methods~\cite{zhang2024text2nerf,hollein2023text2room,ouyang2023text2immersion,wang2024prolificdreamer} produce entangled geometry without object-level separation~\cite{lan20242d}, leading to limited or no editability.
In contrast, DreamScene supports \textbf{automatic layout planning}, ensures \textbf{scene-wide consistency}, achieves \textbf{high visual fidelity}, and enables \textbf{flexible} editing of each individual objects.}}
\label{fig:compare}
\end{figure*}

 Text-to-3D methods~\cite{poole2022dreamfusion,lin2023magic3d,chen2023fantasia3d,liu2023zero,metzer2023latent,huang2023dreamtime,yu2023text,liang2023luciddreamer,tang2023dreamgaussian,li2023sweetdreamer,nichol2022point,jun2023shap} 
 typically use pre-trained 2D text-to-image models~\cite{ramesh2022hierarchical,rombach2022high,saharia2022photorealistic} as prior supervision to create object-centric 3D differentiable representations~\cite{mildenhall2021nerf,park2019deepsdf,kerbl20233d,muller2022instant,shen2021deep} \textcolor{black}{by rendering image from the camera's perspective facing towards the object.} Generating text-to-3D scenes require rendering from preset camera positions outward, capturing the scene from these specific viewpoints. However, as shown in Fig.~\ref{fig:compare}, these text-to-3D generation techniques face several significant obstacles, including: \textcolor{black}{\textbf{1)} A lack of automation, often relying on manual layout design or hardcoded placement trajectories, thereby reducing usability and scalability~\cite{cohen2023set, hollein2023text2room, ouyang2023text2immersion,li2024dreamscene}}; \textbf{2)} Inconsistent 3D visual cues~\cite{hollein2023text2room,zhang2024text2nerf,cohen2023set,po2023compositional,ouyang2023text2immersion,wang2024prolificdreamer,zhang2023scenewiz3d}, with satisfactory outputs restrained to only training camera poses, similar to 360-degree photography, \textcolor{black}{which limits their  applicability in interactive or exploratory tasks within the generated 3D environment.;} 
  \textbf{3)} An inefficient generation process often results in subpar outputs~\cite{zhang2024text2nerf,cohen2023set,lin2023componerf,po2023compositional} and extended completion times~\cite{hollein2023text2room,wang2024prolificdreamer}; 
  \textbf{4)} The inability to distinguish objects from their environments, which obstructs flexible editing on individual components~\cite{hollein2023text2room,zhang2024text2nerf,ouyang2023text2immersion,wang2024prolificdreamer}.
 
 \par \textcolor{black}{To address these limitations, we present \textbf{\textit{\sysname}}, an end-to-end framework  that enables automated, efficient, scene-consistent, and flexibly editable 3D scene generation. Firstly, we perform scene planning by decomposing the scene into structured object-level and environment-level components. Given either an open-ended scene prompt or an interactive dialogue, a GPT-4 agent~\cite{achiam2023gpt} infers detailed information for each object, including its category, real-world size, and descriptive prompt. Based on these results, the agent assigns coarse placements by predicting region-level anchors (e.g., center, side, corner) and inter-object spatial relations (e.g., next to, opposite). We organize these spatial constraints into a hybrid constraint graph, capturing both object-to-object and object-to-scene relationships. To compute a valid layout, we propose a graph-based constraint placement (GCP) algorithm that incrementally assigns position and orientation to each object while avoiding collisions. This yields a physically plausible, semantically consistent object arrangement and provides affine parameters—scaling $s$,translation $t$ and rotation $r$—for each object to be used in downstream generation.}
\par  \textcolor{black}{Secondly, we generate 3D object representations using Formation Pattern Sampling (FPS) guided by descriptive prompts from the planning stage.} Based on the observed patterns in 3D representation formation, FPS utilizes multi-timestep sampling (MTS) to balance semantic information and shape consistency, enabling the rapid generation of high-quality, semantically rich 3D representations. FPS ensures stable generation performance by eliminating redundant internal 3D Gaussians during optimization. And, by employing DDPM~\cite{ho2020denoising} with small timestep sampling and 3D reconstruction techniques~\cite{kerbl20233d}, FPS efficiently generates surfaces with plausible textures from various viewpoints in just \textbf{\textit{tens of seconds}}.

\par \textcolor{black}{Third, we insert the optimized objects into the scene according to the layout inferred in the planning stage, applying their predicted affine transformations to initialize the composition.} We then introduce a progressive three-step camera sampling strategy to create an environment and guarantee 3D consistency. \textbf{1)}, we generate a coarse environment representation by positioning the camera at the center of the scene. \textbf{2)}, we modify ground formation according to the scene type: a) for indoor scenes, by dividing them into regions and choosing a random camera position for rendering; b) for outdoor scenes, by arranging them into concentric circles based on the radius, and sampling camera poses at different circles along the same direction. \textbf{3)}, we solidify the scene through reconstructive generation in FPS, using all camera poses to further refine the scene. This process results in a semantically aligned and visually consistent  scene, mitigating issues such as the multi-headed artifact commonly found in prior text-to-3D scene generation methods~\cite{hollein2023text2room,zhang2024text2nerf,ouyang2023text2immersion,wang2024prolificdreamer} .

\par \textcolor{black}{Finally, \sysname\ supports flexible scene editing through three core operations: \textit{object relocation}, \textit{appearance modification}, and \textit{temporal movement}. Object positions can be adjusted by modifying affine parameters and re-invoking scene planning module. Appearance edits, including shape or texture changes, are enabled via an MTS-based 2D optimization pipeline. For dynamic behaviors, we assign time-dependent transformations to selected objects, allowing them to follow user-specified motion trajectories in 4D scene generation.}

\par \textcolor{black}{This work is an improvement over our ECCV2024 work~\cite{li2024dreamscene}, achieved by substantially extending the method and experiment in the following ways: \textbf{(I)}  We introduce a novel \textit{Scene Planning} module to automatically generate structured, layout-aware 3D scenes. Instead of manually defining object placements, we leverage GPT-4~\cite{achiam2023gpt} as an agent to infer object categories, physical dimensions, and spatial constraints from either direct descriptions or multi-turn dialogues. A hybrid constraint graph is constructed to represent object-to-object and object-to-scene relations, and a graph-based constraint placement (GCP) algorithm assigns valid, collision-free positions and orientations.  The inferred layout aligns with common sense and physical feasibility, serving as a strong prior for downstream environment generation and helping prevent artifacts such as multi-headed scenes. (Sec.~\ref{sec:Scene Planning}).~\textbf{(II)} We develop a flexible editing framework for post-hoc scene control, supporting: (a) object relocation via affine updates and planning re-execution; (b) appearance editing by adapting MTS-based 2D diffusion to our 3D pipeline; and (c) motion editing through time-varying transformations for dynamic 4D scene composition (Sec.~\ref{sec:scene-editing}). ~\textbf{(III)} We provide a theoretical explanation of Multi-Timestep Sampling (MTS), showing its connection to 2D editing frameworks (Sec.I in Supp.). ~\textbf{(IV)} We provide a more comprehensive analysis and evaluation of current text-to-3D scene generation methods. This includes an expanded discussion of a technical comparison between \sysname\ and prior approaches and layout generation strategies  (Sec.\ref{sec:text-to-3d scene}), along with additional camera sampling details (Sec.\ref{sec:qualitative results}) and extended qualitative and quantitative experiments on layout generation, scene generation quality, scene editing and camera sampling  (Sec.~\ref{sec:qualitative results}, Sec.~\ref{sec:quantitative-results}, Sec.~\ref{sec:scene-editing-result}, Sec.~\ref{sec:ablations}).}

\section{Related Work}

\subsection{Differentiable 3D Representation}
Utilizing differentiable approaches such as NeRF~\cite{mildenhall2021nerf,barron2021mip}, SDF~\cite{park2019deepsdf,shen2021deep}, and 3D Gaussian Splatting~\cite{kerbl20233d}, it becomes possible to represent, manipulate, and render 3D objects and scenes effectively. These kinds of representations work well with optimization algorithms like gradient descent, making it feasible to automatically adjust the parameters of 3D representations by minimizing loss. A notable recent development~\cite{kerbl20233d} involves the use of differentiable 3D Gaussians to model 3D scenes, which has resulted in exceptional real-time rendering performance through the splatting technique. In comparison to implicit representations~\cite{mildenhall2021nerf,barron2021mip,muller2022instant}, 3D Gaussians present a more explicit framework that eases the integration of multiple scenes. Consequently, we select 3D Gaussians for their straightforward, explicit representation and the simplicity associated with merging scenes.
\subsection{Text-to-3D Generation}
Currently, the main approaches to generating 3D representations in text-to-3D tasks involve either direct methods~\cite{nichol2022point,jun2023shap,shi2023mvdream} or distillation from pre-trained 2D text-to-image models~\cite{poole2022dreamfusion,lin2023magic3d,chen2023fantasia3d,li2024text}. Direct techniques require annotated 3D datasets for quick generation, but they frequently face issues such as lower quality and increased GPU demands, often acting as initial stages for distillation methods~\cite{liang2023luciddreamer,yi2023gaussiandreamer}. For instance, Point-E~\cite{nichol2022point} creates an image by employing a diffusion model based on text, which is subsequently transformed into a point cloud. Conversely, Shap-E~\cite{jun2023shap} links 3D assets to implicit function parameters using an encoder and trains the  diffusion model  based on these parameters with conditions.
\par The prevailing approach in the field has become the distillation of 3D representations from pre-trained 2D text-to-image diffusion models~\cite{poole2022dreamfusion,lin2023magic3d,chen2023fantasia3d,metzer2023latent,huang2023dreamtime,yu2023text}. A pioneer, DreamFusion~\cite{poole2022dreamfusion}, blazed a trail by introducing Score Distillation Sampling (SDS), ensuring that images rendered from multiple viewpoints align with the distribution of 2D text-to-image models~\cite{ramesh2022hierarchical,rombach2022high,saharia2022photorealistic}. Subsequent advancements~\cite{lin2023magic3d,chen2023fantasia3d,liu2023zero,metzer2023latent,huang2023dreamtime,yu2023text,liang2023luciddreamer,tang2023dreamgaussian,li2023sweetdreamer,nichol2022point,jun2023shap} have built upon this, refining 3D generation in terms of quality, speed, and diversity. For instance, LucidDreamer~\cite{liang2023luciddreamer} employs DDIM inversion~\cite{mokady2023null,hertz2022prompt} to ensure 3D consistency during the object generation process, while DreamTime~\cite{huang2023dreamtime} hastens the generation convergence via monotonically non-increasing sampling of timestep $t$ in a 2D text-to-image model. Drawing inspiration from these pioneering works, our method offers a more efficient route to generate high-quality and semantically rich 3D representations.
\vspace{-8pt}
\subsection{Text-to-3D Scene Generation Methods}

\label{sec:text-to-3d scene}
\textcolor{black}{Contemporary text-to-3D scene generation techniques, as depicted in Fig.\ref{fig:compare}, encounter considerable constraints. We can classify these methods into three categories: Inpainting-based methods\cite{zhang2024text2nerf,hollein2023text2room,ouyang2023text2immersion}, Combination-based methods~\cite{cohen2023set,zhang2023scenewiz3d}, and Layout generation methods~\cite{vilesov2023cg3d,zhou2024gala3d,lin2023componerf}.}

\noindent \textcolor{black}{\textbf{Inpainting-based methods}~\cite{zhang2024text2nerf,hollein2023text2room,ouyang2023text2immersion} utilize text-to-image inpainting techniques for generating scenes and currently serve as the main approach for scene generation. These methods initiate an image, partially mask it to represent a different viewpoint, and then employ pretrained image inpainting models like Stable Diffusion~\cite{rombach2022high} along with depth estimation to reconstruct the concealed parts of the image and infer their depths. The entire scene is iteratively composed through depth and image alignment. Although these methods can yield visually appealing results at specific camera positions(e.g., the scene's center) during the generation process, their visible range faces substantial limitations. Exploring beyond the predefined camera areas used during generation leads to scene deterioration, as illustrated in Fig.~\ref{fig:consistency_outdoor} and Fig.~\ref{fig:consistency_indoor}, highlighting a lack of 3D consistency throughout the scene. More critically, generated scenes exhibit a "multi-head" issue, similar to the multiple heads appearing in object generation methods~\cite{poole2022dreamfusion,lin2023magic3d,chen2023fantasia3d}. In the scene context, this translates to multiple identical objects appearing in various directions, such as several sofas facing different directions in a living room. By employing a carefully devised camera sampling strategy and pre-positioning objects in the scene to guide the generation of the surrounding environment, \sysname\ attains scene-wide consistency and reasonable environmental content creation.}
\par \noindent \textcolor{black}{\textbf{Combination-based methods}~\cite{cohen2023set,zhang2023scenewiz3d} leverage an assembly approach for scene construction.  They grapple with issues such as subpar generation quality and sluggish training rates. In addition, \cite{zhang2023scenewiz3d} makes use of multiple 3D representations (such as NeRF$+$DMTet) for integrating objects and scenes, which heightens the intricacy of scene representation and restricts the number of objects that can be incorporated within the scene (2-3 objects), thereby impacting their utility. Conversely, \sysname's FPS method can swiftly generate high-quality 3D content, by using a solitary 3D representation to assemble the entire scene, which allows for the inclusion of over 20 objects within the scene.}
\par \noindent \textcolor{black}{\textbf{Layout generation methods} adopt diverse strategies. Methods~\cite{vilesov2023cg3d,zhou2024gala3d,lin2023componerf,zhou2024layout,nath2025decompdreamer}, such as CG3D~\cite{vilesov2023cg3d}, typically rely on structured scene prompts and optimize layout parameters via image-based supervision. They focus primarily on the logical assembly of a small set of objects while neglecting broader environmental context, resulting in basic arrangements rather than comprehensive scenes. These methods also struggle with occlusion and local minima as layout complexity increases. CC3D~\cite{bahmani2023cc3d} generates layout-conditioned 3D scenes by back-projecting 2D diffusion outputs into NeRF fields, but requires the layout to be explicitly provided. BerfScene~\cite{zhang2024berfscene} reconstructs fused volumetric 3D scenes from single images without object-level structure or layout control. ATISS~\cite{paschalidou2021atiss} autoregressively generates indoor layouts from structured priors using Transformers, yet remains limited to closed indoor domains and requires floorplan input. In contrast, \sysname\ supports open-ended prompts or dialogues and generates diverse and reasonable layouts instead of a single fixed arrangement. Furthermore,  unlike Scene-LLM~\cite{fu2024scene} and 3D-LLM~\cite{hong20233d}, which focus on understanding or interacting with existing 3D scenes/layouts and rely heavily on limited indoor datasets for supervision~\cite{wang2024root}, our approach generates complex 3D scenes entirely from scratch. By leveraging GPT-4's~\cite{achiam2023gpt} broad knowledge of the physical world, \sysname\ supports open-domain scene generation beyond the constraints of pre-collected 3D data.}

\par \textcolor{black}{ \sysname\ exhibits a significant edge by autonomously generating 3D scenes with efficiency, consistency, and flexibility, surpassing prior methods. }

\vspace{-5pt}
\section{preliminary} 

\par \noindent \textbf{Diffusion Models}~\cite{ho2020denoising,song2020denoising} facilitate the generation of data 
 $x$($x\sim p(x)$)  by approximating the gradients of log probability density functions, represented as $\nabla_x \log p_{data}(x)$. During training, noise is progressively added to the input 
$x$ across $t$ distinct steps:
\begin{equation}
    x_t = \sqrt{\bar{\alpha_t}}x + \sqrt{1-\bar{\alpha_t}}\epsilon , 
    \label{Eq:ddpm}
\end{equation}
where $\bar{\alpha_t}$ denotes a predetermined coefficient and $ \epsilon$, representing noise, is drawn from a normal distribution $\mathcal{N}(0,I)$. The noise prediction network $\phi$ then optimized by reducing the prediction loss $\mathcal{L}_t$:
\begin{equation}
\mathcal{L}_t = \mathbb{E}_{x, \epsilon \sim \mathcal{N}(0, I)}\left[\left\lVert \epsilon_{\phi}(x_t, t) - \epsilon \right\rVert^2\right] .
\end{equation}
In the sampling phase, the method deduces 
$x$  using both the noisy input and its estimated noise $\epsilon_{\phi}(z_t, t)$.

\begin{figure*}[t]
\centering
\includegraphics[width=\textwidth]{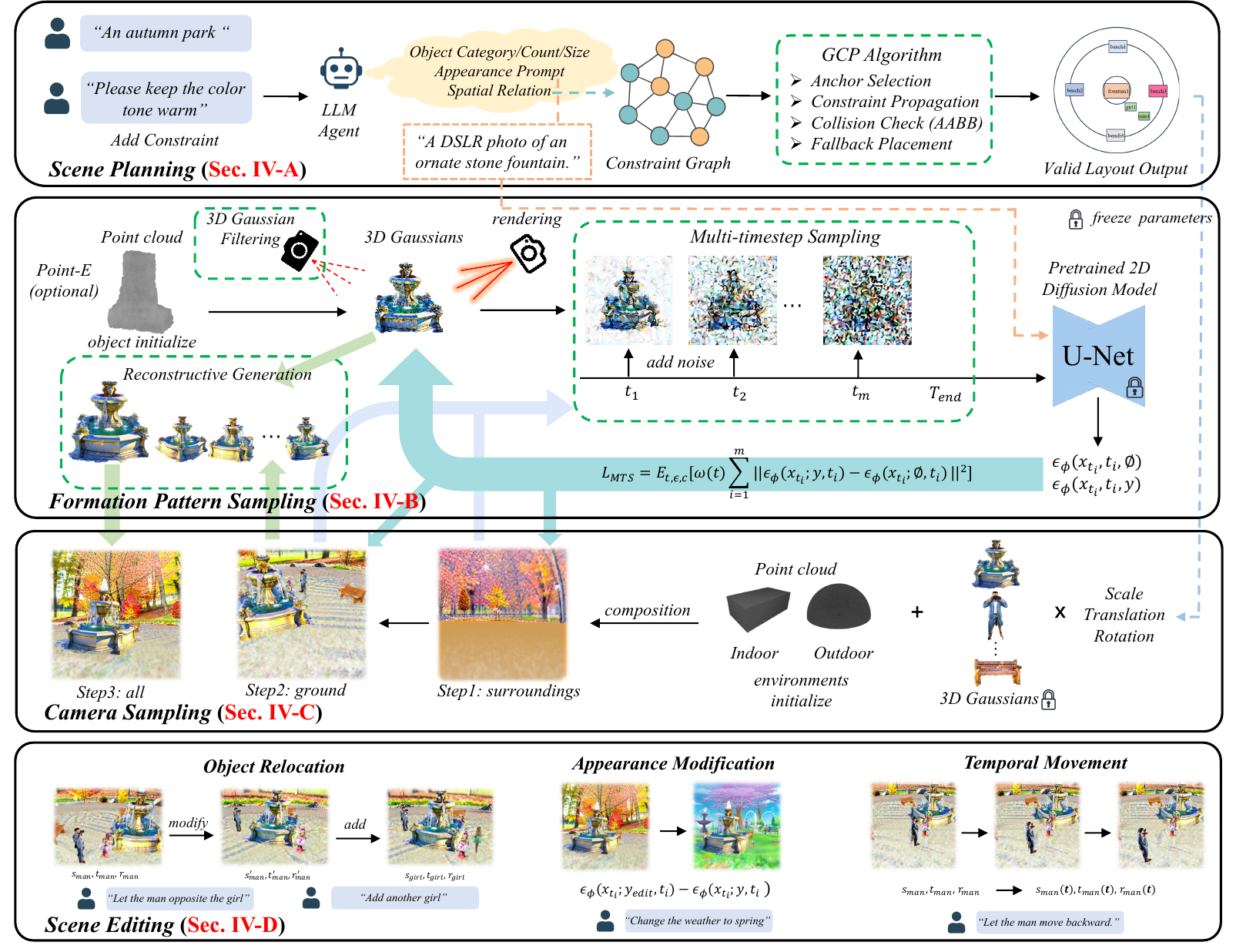}
\caption{ \textcolor{black}{Our framework enables automatic 3D scene generation from natural language, supporting both direct descriptions and interactive dialogues. A GPT-4 agent first performs scene decomposition by inferring object semantics, layout constraints, and spatial relations, and constructs a constraint graph to plan collision-free object placements. Each object is generated using Formation Pattern Sampling (FPS), which integrates multi-timestep sampling, 3D Gaussian filtering, and reconstructive generation. These objects are placed into the global scene using predicted affine transformations. We then apply a three-stage camera sampling strategy to optimize the environment and ensure scene-wide consistency. \sysname\ also supports structure-aware scene editing, including object relocation, appearance modification, and 4D editing.  }
}
\label{fig:overview}
\end{figure*}

 \par \noindent \textbf{Score Distillation Sampling (SDS)} technique, introduced by DreamFusion~\cite{poole2022dreamfusion}, aims to distill 3D representations from a pre-trained 2D text-to-image diffusion model. The approach involves a differentiable 3D representation, parameterized by $\theta$ and a rendering function, $g$. For a specified camera pose $c$, the image $x$ is rendered as $x = g(\theta,c)$. Subsequently, SDS employs a 2D diffusion model $\phi$ with fixed parameters to distill $\theta$ by:
\begin{equation}
    \nabla_{\theta}\mathcal{L}_{\text{SDS}}(\theta) = \mathbb{E}_{t,\epsilon,c}\left[w(t)(\epsilon_{\phi}(x_t; y, t) - \epsilon)\frac{\partial g(\theta,c)}{\partial \theta}\right],  
\label{eq:SDS}
\end{equation}
where $w(t)$ serves as a weighting function that adjusts based on the timesteps $t$ and 
$y$ represents the text embedding derived from the input prompt.
\par \noindent \textbf{Classifier Score Distillation (CSD)}~\cite{yu2023text} is a variation of Score Distillation Sampling(SDS) and takes its cue from Classifier-Free Guidance (CFG)~\cite{ho2022classifier}.  This technique differentiates the noise variance in SDS into two components:  
the generation prior, noted as $\epsilon_{\phi}(x_t; y, t) - \epsilon$, and the classifier score, noted as $\epsilon_{\phi}(x_t; y, t) - \epsilon_{\phi}(x_t; \emptyset, t)$, $\emptyset$ represents the empty prompt. This approach suggests that the classifier score is robust enough to facilitate text-to-3D translation, and it is outlined as follows:
\begin{equation}
\resizebox{0.90\linewidth}{!}{
    $\nabla_{\theta}\mathcal{L}_{\text{CSD}}(\theta) = \mathbb{E}_{t,\epsilon,c}\left[w(t)(\epsilon_{\phi}(x_t; y, t) - \epsilon_{\phi}(x_t; \emptyset, t))\frac{\partial g(\theta,c)}{\partial \theta}\right]$.}
\label{eq:CSD}
\end{equation}

\par \noindent \textcolor{black}{\textbf{DreamTime}~\cite{huang2023dreamtime} is an SDS-based~\cite{poole2022dreamfusion} time sampling strategy that posits that sampling larger timestep $t$  at the beginning of the iteration and smaller timestep $t$ later can accelerate convergence of 3D model generation. Therefore, it introduces a monotonically non-increasing sampling of timestep $t$. Specifically, it defines a function $W(t)$ for $t$, where larger values indicate that the current $t$ is significant and should be sampled flatly, while smaller values suggest a steep sampling.
\begin{equation}
    W(t) = \frac{1}{Z} \sqrt{\frac{1 - \alpha_t}{\bar{\alpha}_t}} e^{-\frac{(t-m)^2}{2s^2}},
\label{eq:dreamtime}
\end{equation}
where $Z = \sum_{t=1}^T\sqrt{\frac{1 - \alpha_t}{\bar{\alpha}_t}} e^{-\frac{(t-m)^2}{2s^2}}$, $s$ and $m$ are  hyper-parameters.}
\textcolor{black}{In fact, such timestep $t$ sampling can indeed increase the model's convergence speed, but it has a little impact on the improvement of 3D representation quality. }

\par  \noindent \textbf{3D Gaussian Splatting}~\cite{kerbl20233d, chen2024survey} represents a novel approach in 3D reconstruction. It involves a 3D Gaussian defined by a comprehensive 3D covariance matrix $\Sigma$ which is established in the world space and centered at a specific point, known as the mean $\mu$:
\begin{equation}
G(\mathbf{x}) = e^{-\frac{1}{2}\mathbf{x}^T\Sigma^{-1}\mathbf{x}},
\end{equation}
spherical harmonics(SH) coefficients and opacity $\alpha$. By implementing interlaced optimization and density control of these 3D Gaussians, particularly through the tuning of the anisotropic covariance, we can get highly accurate reconstruction representations. Additionally, a tile-based rendering strategy is utilized to facilitate efficient anisotropic splatting, which not only speeds up the training process but also enables real-time rendering capabilities.




\section{method}

\begin{figure*}
\centering
\includegraphics[width=\textwidth]{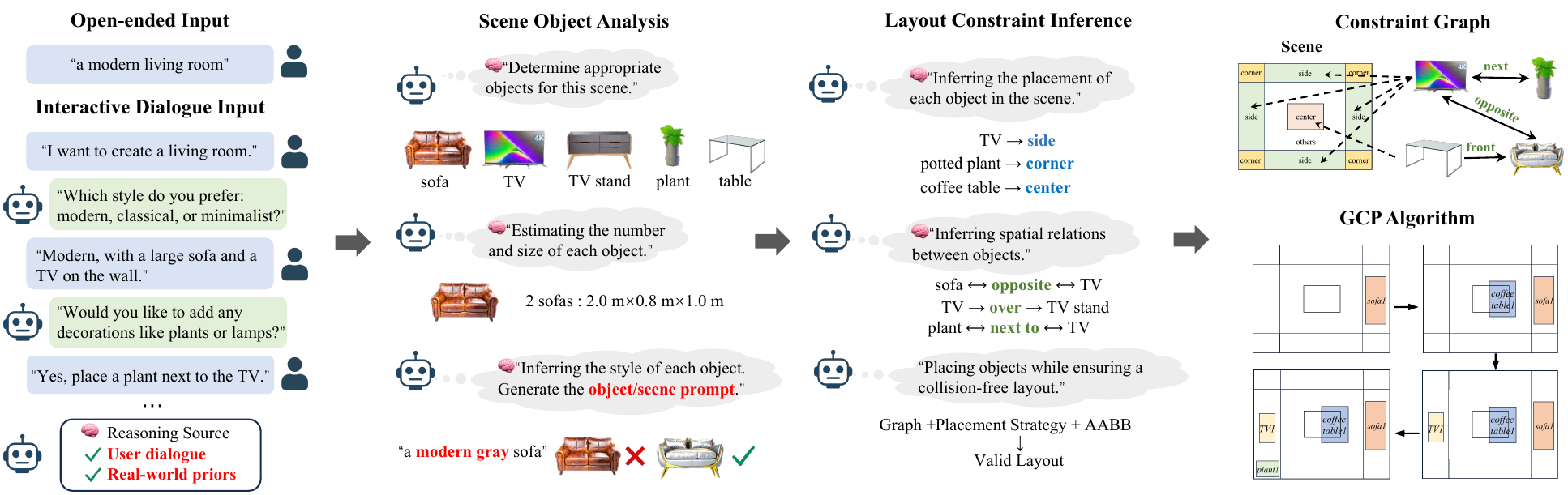}
\caption{ \textcolor{black}{Overview of the Scene Planning process. Given either an open-ended prompt or an interactive dialogue, a GPT-4 agent infers object categories, real-world sizes, textual prompts, spatial placements, and inter-object relations. These constraints are used to plan the layout through a constraint graph and GCP algorithm. The resulting arrangement provides a physically plausible and semantically coherent layout that supports environment generation.}}
\label{fig:scene-planning}
\end{figure*}

\textcolor{black}{We present an end-to-end framework \sysname\ for \textbf{automatic} 3D scene generation from natural language inputs, supporting both \textit{direct descriptions} and \textit{interactive dialogues}. The system jointly infers object/scene semantics, spatial layout, and stylistic consistency, and produces \textbf{high-quality} scenes with \textbf{scene-wide consistency} and \textbf{flexibility} for editing.}
\par \textcolor{black}{The generation process begins with a Scene Planning module, where a GPT-4 agent infers object categories, real-world sizes, detailed textual descriptions $y$, spatial relations, and region-level placement anchors. It constructs a hybrid constraint graph and applies a graph-based constraint placement (GCP) algorithm to produce a structured, collision-free object arrangement, from which we derive the affine transformation parameters for each object, including scaling $s$, rotation $r$, and translation $t$. Each object is subsequently generated using Formation Pattern Sampling (FPS), conditioned on the corresponding description $y$. FPS incorporates multi-timestep sampling (MTS), 3D Gaussian filtering, and reconstructive generation, enabling the rapid synthesis of high-quality 3D content using a minimal number of Gaussians. For environment generation, we first initialize cuboid 3D Gaussians to represent indoor elements such as walls, floors, and ceilings, and hemispherical Gaussians for outdoor backgrounds like ground and distant surroundings. We then place each of the $N$ generated objects into the global scene coordinate system using the predicted affine transformations:
\begin{equation}
world(x_i) = r_i\cdot s_i \cdot o_i(x) + t_i, i=1,...,N,
\label{eq:coordinate}
\end{equation}
where $x_i$ denotes the  coordinates of all 3D Gaussians belonging to object $i$. Finally, we implement a camera sampling strategy to guide the three-stage optimization of the environment, ensuring scene-wide 3D consistency and mitigating common scene-level issues such as “multi-headed” layouts, where identical objects (e.g., sofas) appear redundantly across multiple directions. Our framework further supports structure-aware 3D scene editing, including object-level relocation via affine transformation updates and flexible modification of scene content using our editing optimization algorithm Additionally, we extend the editing capability to the temporal dimension, enabling \textbf{4D scene} editing with controllable object motion over time.}

\vspace{-8pt}
\subsection{Scene Planning}
\label{sec:Scene Planning}


\textcolor{black}{To support the goal of \textit{\sysname}, which aims to generate diverse and open-domain 3D scenes, we adopt GPT-4~\cite{achiam2023gpt} as the core reasoning agent for scene planning. Unlike methods~\cite{fu2024scene, hong20233d } constrained by specific indoor datasets, our approach requires the ability to infer rich world knowledge, resolve spatial relationships, and generate layout-aware prompts across a wide range of scenes.}
\par \textcolor{black}{As illustrated in Fig.~\ref{fig:scene-planning}, user input can take the form of either an open-ended description (e.g., “a modern living room”) or an interactive dialogue where the agent proactively queries preferences, such as style or functional constraints. These interactions form a contextual history that, together with commonsense priors, guides the generation of all downstream prompts. Specifically, we prepend each GPT-4 query with the phrase \textit{“Based on the user history dialogue and real-world priors”} to   ensure that the generated descriptions and layouts satisfy user intent and adhere to real-world spatial and functional constraints.}

\subsubsection{Scene Object Analysis}
\textcolor{black}{Based on the user dialogue and scene intent, the GPT-4 agent first infers a list of candidate objects that are likely to appear in the scene as shown in Fig.~\ref{fig:scene-planning}. For each object, it predicts the category, count, real-world size, and a fine-grained textual description \( y_i \). These descriptions capture both functional roles (e.g., “a low wooden coffee table”) and stylistic attributes (e.g., “a modern gray sofa”) inferred from the dialogue and high-level scene goal. To guide this process, we design a structured prompt that instructs GPT-4 to act as a professional scene designer and return object-level information in \textbf{JSON} format. The output includes the number of instances, physical dimensions (in meters), and a descriptive caption starting with “A DSLR photo of” to encourage photorealistic generation. The full prompt template and an example are provided in the supplementary material. To reduce computational cost, we generate one object instance per category and replicate it according to the predicted count. To introduce diversity, these replicas can be associated with slightly varied prompts, allowing the system to produce stylistic variations of the same object type without regenerating the geometry from scratch.}

\subsubsection{Layout Constraint Inference}
\textcolor{black}{To obtain plausible and controllable spatial layouts, we prompt the GPT-4 agent to infer layout constraints from the object list $\mathcal{O} = \{o_1, o_2, ..., o_N\}$ . This includes two levels of constraint generation: (1) object-to-scene region anchors $\mathcal{A}_i$ and (2) object-to-object spatial relations. These constraints serve as soft guidance for downstream layout search, enabling position reasoning without relying on supervised 3D layout annotations.}
\par \textcolor{black}{For region anchoring $\mathcal{A}$ , we divide the scene into coarse semantic zones. In indoor scenes, these include \textit{center}, \textit{side}, \textit{corner}, and \textit{others}, while in outdoor scenes we exclude the \textit{corner} zone due to the lack of enclosing structure. The GPT-4 agent is prompted to assign each object to an appropriate zone based on its name, function, and contextual relevance to the scene. For example, coffee tables are typically centered in a living room, while plants or shelves may be placed at the periphery or in corners. A visual illustration of the region definitions for an indoor scene is shown in the top-right corner of Fig.~\ref{fig:scene-planning}. To enhance the plausibility of object placements, we further query GPT-4 to infer pairwise spatial relations among objects using a limited relation set: \textit{left}, \textit{right}, \textit{front}, \textit{back}, \textit{over}, \textit{under}, \textit{next} and  \textit{opposite}. These relations are simple yet expressive, capturing typical scene configurations such as “TV opposite sofa” or “lamp next to table.” The prompt templates used to generate both region anchors and object relations are provided in the supplementary material.}

\subsubsection{Constraint-based Layout Generation}
\textcolor{black}{Given the layout constraints inferred by GPT-4, we construct a constraint graph $\mathcal{G}$ where nodes $\mathcal{V}$ represent objects and edges $\mathcal{E}$ encode pairwise spatial relations. To realize a plausible and collision-free layout, we propose a graph-based constraint placement (GCP) algorithm, as shown in Algorithm~\ref{alg:layout}, which incrementally assigns object positions and rotations within the scene.}
\begin{algorithm}[t]
\caption{Graph-based Constraint Placement}
\label{alg:layout}
\begin{algorithmic}[1]
\Require 
    Object set $\mathcal{O} = \{o_1, o_2, ..., o_N\}$ \\
    Constraint graph $\mathcal{G} = (\mathcal{V}, \mathcal{E})$ with spatial relations \\
    Anchor region $\mathcal{A}_i$ and real-world size for each object $o_i$
\Ensure 
    Position/Translation $\{t_i\}$, rotation $\{r_i\}$, and scaling $\{s_i\}$ for all objects

\State Compute scaling factor $s_i$ as the ratio between real-world size and model size for each $o_i$
\State Initialize candidate positions $\mathcal{C}_i$ from anchor region $\mathcal{A}_i$
\State Select anchor object $o_a$ (e.g., most connected or central)
\State Estimate initial rotation $r_a$ based on anchor orientation
\State Initialize placement queue $\mathcal{Q} \leftarrow [o_a]$ and mark $o_a$ as placed

\While{$\mathcal{Q}$ not empty}
    \State Pop object $o_i$ with known position $t_i$ and rotation $r_i$
    \For{each unplaced neighbor $o_j$ of $o_i$ in $\mathcal{G}$}
        \State Retrieve spatial relation $r_{ij}$ from $\mathcal{G}$
        \State Use $t_i$ and $r_i$ to infer $o_j$'s directional constraint  
        \State Filter $\mathcal{C}_j$ to satisfy $r_{ij}$ and avoid AABB collisions
        \If{$\mathcal{C}_j$ is not empty}
            \State Select $t_j \in \mathcal{C}_j$, infer $r_j$ accordingly
            \State Mark $o_j$ as placed, enqueue $o_j$ into $\mathcal{Q}$
        \Else
            \State Defer placement of $o_j$
        \EndIf
    \EndFor
\EndWhile

\For{each unplaced object $o_k$}
    \State Assign fallback position $t_k$ and estimate $r_k$ heuristically
\EndFor

\State \Return $\{t_i, r_i, s_i\}$ for all $o_i$
\end{algorithmic}
\end{algorithm}
\textcolor{black}{We begin by computing the scaling factor $s_i$ for each object $o_i$, defined as the ratio between its real-world dimensions and the default size of its generated 3D model. This ensures correct physical scale in the scene and provides a reliable basis for collision checking. Based on the region anchors $\mathcal{A}_i$, we sample a set of candidate positions $\mathcal{C}_i$ for each object on a discretized spatial grid. We then select an anchor object $o_a$, typically the one with the most relational connections, and initialize its rotation $r_a$ according to its anchor direction (e.g., facing the center if placed at the boundary). Object rotations serve as the spatial reference frame to resolve directional constraints such as \textit{left}, \textit{front}, or \textit{opposite}. Starting from $o_a$, we propagate placements through the graph in a breadth-first manner. For each neighboring object $o_j$, we use the relation $r_{ij}$ and the current object’s pose to filter valid candidates from $\mathcal{C}_j$, retaining only those that satisfy the directional constraint and avoid AABB collision. If such candidates exist, we assign one based on simple heuristics such as proximity to anchor or alignment with room center, and infer $r_j$ accordingly; otherwise, we defer placement. After traversal, deferred objects are assigned fallback positions, and their rotation is estimated based on nearby anchors or previously placed objects. The final output of this process is a complete layout specification $\{t_i, r_i, s_i\}$ for each object.}

\textcolor{black}{The resulting layout aligns with real-world spatial logic and provides a strong structural prior for downstream environment generation, effectively mitigating multi-headed arrangements in the scene.}

\subsection{Formation Pattern Sampling}
\label{sec:Formation Pattern Sampling}
We have enhanced and expanded the concept of employing monotonically non-increasing sampling of timestep
$t$ in DreamTime~\cite{huang2023dreamtime}. Our research indicates that developing high-quality, semantically rich 3D representations greatly benefits from \textbf{\textit{integrating information across multiple timesteps}} at each iteration of a pre-trained text-to-image diffusion model. This approach stands in contrast to other methods using Score Distillation Sampling~\cite{poole2022dreamfusion}(SDS), which typically rely on information from a single timestep during each iteration.
\textcolor{black}{In the optimization's early to mid stages, which target the initial shaping of forms, a decremental time window $T_{end}$ is implemented, linearly reducing through iterations. This window is segmented into $m$ intervals; within each $t$ is randomly selected for gradient aggregation. Although this method quickly produces rich semantic 3D representations, it may also generate unnecessary massive 3D Gaussians. To counter this, we employ 3D Gaussian filtering, selectively sampling critical surface Gaussians only. In later optimization stages, to make the surface textures of representations more plausible, we sample $t$ from a range between $0$ and $200$ using 3D reconstruction techniques~\cite{kerbl20233d} to expedite this process. 
Since this method for generating 3D representations follows the patterns of 3D model development, sampling different time steps $t$ in various iterations and targeting 3D Gaussians on the model's surface, we aptly named it \textbf{\textit{Formation Pattern Sampling}} (FPS). }

\par To capture the varied information offered by the 2D text-to-image diffusion model across timestep $t$ ranging from $0$ to $1000$, we utilize pseudo-Ground-Truth(pseudo-GT) images generated from a single denoising step within LucidDreamer~\cite{liang2023luciddreamer}. By introducing noise across $t$ timestep into the images 
$x_0$ to generate $x_t$
 , we calculate the pseudo-GT $\hat{x}_{0}^t$ using the following equation:

\begin{equation}
\hat{x}_{0}^t = \frac{x_{t} - \sqrt{1 - \bar{\alpha}_t} \epsilon_{\phi}(x_{t}; y, t)}{\sqrt{\bar{\alpha}_t}}.
\label{Eq:pseudoGT}
\end{equation}
\subsubsection{Multi-timestep Sampling}
As illustrated in Fig.~\ref{fig:Formation Pattern Sampling} (a), we observe that at smaller timestep $t$, the 2D diffusion model produces detailed and realistic surface textures that align well with the current 3D shape, but lack comprehensive semantic information from the prompt $y$. Conversely, at a larger timestep $t$, the model provide richer semantic details, though these may not conform to the existing 3D shape(discrepancies in the orientation of the man, the color of the chair, or the direction of a cooker between timestep $600$ and $800$).

\par To address this, we suggest blending information from multiple timesteps in each iteration of a 2D diffusion model. This integration aims to maintain shape accuracy while enhancing semantic information. For example, during the $300$-th iteration for the man in Fig.~\ref{fig:Formation Pattern Sampling} (a), we utilize timesteps $200$ to $400$ for shape accuracy, while timesteps $400$ to $600$ and $600$ to $800$ enrich the semantic context. However, by the $1000$-th iteration for the cooker, we note that the shape already encapsulates sufficient semantic details, and incorporating further information from a larger timestep might detract from the optimization process. So the timestep $t$ for 
$i$-th sample can be described as follows: 
\begin{equation}
    t_i = T_{end}^{iter}\cdot random(\frac{i-1}{m},\frac{i}{m}),i=1,...,m,
\end{equation}
where $T_{end}$ represents a linearly decreasing time window, akin to the approach used in DreamTime~\cite{huang2023dreamtime}, 
$iter$  indicates the current iteration, and 
 $m$ specifies the number of intervals.
\textcolor{black}{Some studies~\cite{wu2024consistent3d,liang2023luciddreamer} have found that using ordinary differential equation(ODE) processes in sampling can ensure a certain level of consistency. Naturally, combining our multi-step consideration, we use DDIM Inversion to calculate $x_{t_i}$ between $t_1$ and $t_m$:}
\begin{equation}
\resizebox{0.89\linewidth}{!}{
        $x_{t_{i+1}} = \sqrt{\bar{\alpha}_{t_{i+1}}} \frac{x_{t_i} - \sqrt{1-\bar{\alpha}_{t_i}}\epsilon_{\phi}(x_{t_i};\emptyset,t_i)}{ \sqrt{\bar{\alpha}_{t_i}}} + \sqrt{1-\bar{\alpha}_{t_{i+1}}}\epsilon_{\phi}(x_{t_i};\emptyset,t_i)$,}
\label{eq:ddim_inversion}
\end{equation}
where $\emptyset$ represents the empty prompt.
\par Therefore, the combination of MTS and CSD~\cite{yu2023text} method can be articulated as follows:
\begin{equation}
\resizebox{0.89\linewidth}{!}{
 $\nabla_{\theta}\mathcal{L}_{\text{MTS}}(\theta) = \\ \mathbb{E}_{t,\epsilon,c}  \left[\sum\limits_{i=1}^{m} w(t_i)(\epsilon_{\phi}(x_{t_i}; y, t_i) - 
 \epsilon_{\phi}(x_{t_i}; \emptyset, t_i))\frac{\partial g(\theta,c)}{\partial \theta} \right]$.}
\label{eq:MTS}
\end{equation}
\par \textcolor{black}{Although MTS is initially motivated by empirical observations across diffusion timesteps, we further provide a theoretical explanation by linking it to trajectory alignment in 2D editing methods~\cite{mokady2023null,dong2023prompt}. In addition, we reduce the estimation error within MTS, which leads to improved generation quality as shown in Fig.~\ref{fig:improve}. Details are presented in the supplementary material. Details are presented in the supplementary material.}

\begin{figure}
    \centering
    \includegraphics[width=\linewidth]{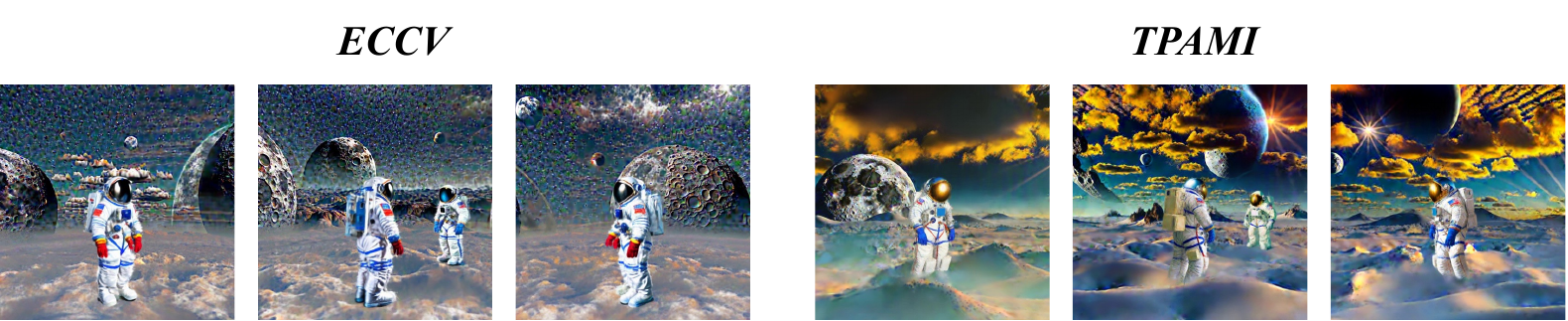}
    \caption{\textcolor{black}{Comparison of the generation quality between the ECCV version and the TPAMI version of \sysname.}}
    \label{fig:improve}
\end{figure}
\subsubsection{3D Gaussian Filtering}
\begin{figure}
    \centering
    \includegraphics[width=\linewidth]{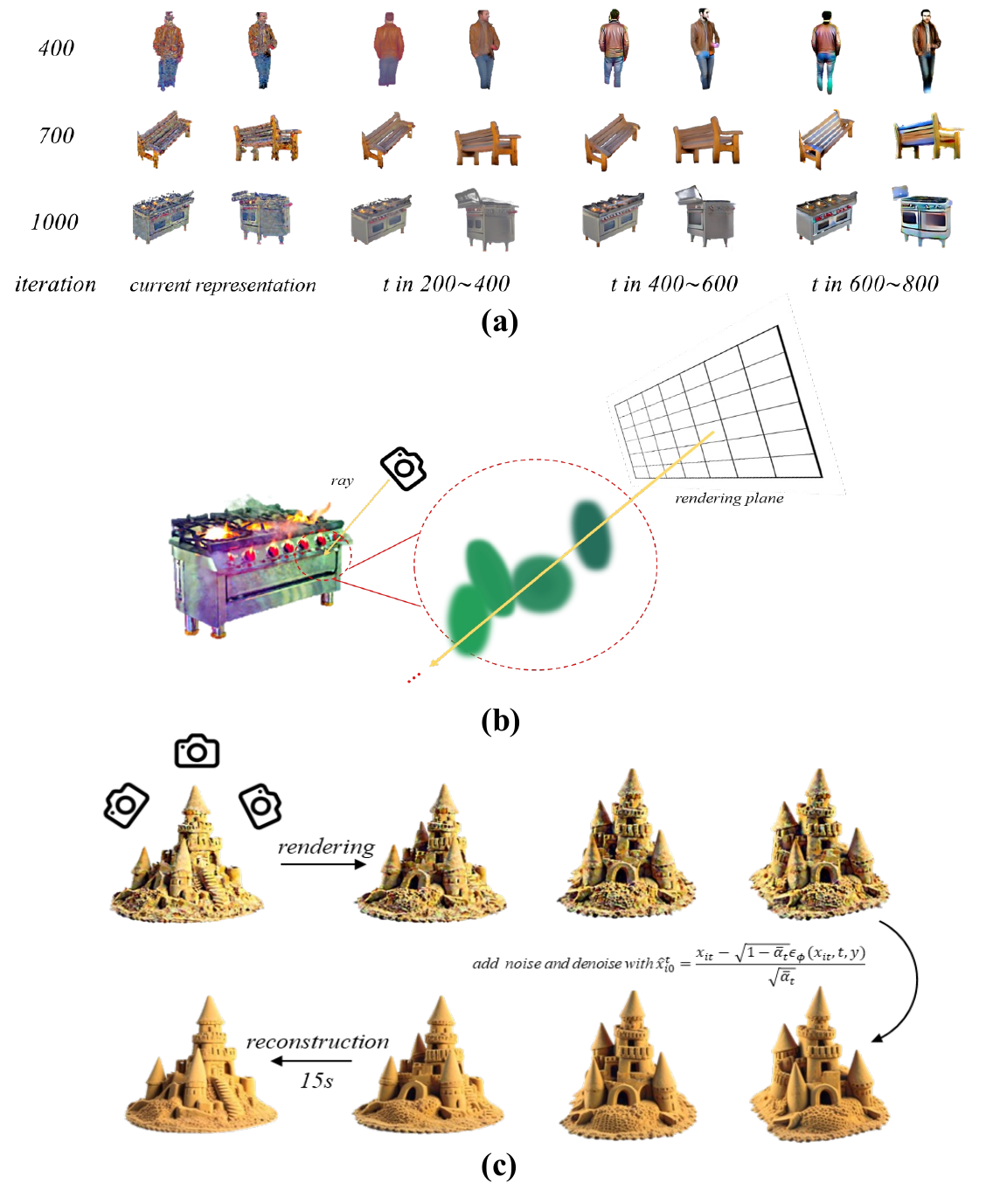}
    \caption{Formation Pattern Sampling. (a)\textbf{Multi-timestep Sampling.} At varying timesteps, the 2D text-to-image diffusion model provides different information(represented by the pseudo-GT $\hat{x}_0^t)$ obtained from $x_t$ in a single-step by Eq.~\ref{Eq:pseudoGT} in LucidDreamer~\cite{liang2023luciddreamer}. (b)\textbf{3D Gaussian Filtering}. 3D Gaussians that are located closer to the rendering plane and possess larger volumes make a greater contribution to the rendering process. (c)\textbf{Reconstructive Generation.} During the later stages of optimization, generation can be directly accomplished using reconstruction based on denoised images, leading to 3D representations with refined and plausible  textures.}
    \label{fig:Formation Pattern Sampling}
\end{figure}

Excessive 3D Gaussians can impede the optimization process. Unlike traditional methods~\cite{fan2023lightgaussian,lee2023compact} that use ground truth images to filter reconstructed 3D Gaussians, our strategy requires filtering to be integrated into the optimization phase. Regarding rendering, 3D Gaussians located nearer to the rendering plane have a more pronounced effect, for which a specialized score function is utilized to evaluate their impact. For 3D Gaussians along the rendering ray $r_j$, their contributions are assessed based on the inverse square of their distance to the rendering plane, factoring in the 3D Gaussians' volume. This technique prioritizes 3D Gaussians that are both closer to the rendering plane and have a larger volume, as illustrated in Fig.~\ref{fig:Formation Pattern Sampling} (b). By scoring various viewpoints, we can effectively discard 3D Gaussians that do not meet a set threshold.
\begin{equation}
Score(i) = \sum_{j=1}^{H\times W \times M} \frac{V(i)}{D(r_j,i)^2\times maxV(r_j)},
\end{equation}
where $H$ and $W$ indicate the height and width of the rendered image, respectively, $M$ represents the number of rendered images, $V(i)$ is the volume of the $i$-th 3D Gaussian(calculated using the covariance matrix), $maxV(r_j)$ is the maximum volume of the 3D Gaussians on $r_j$, and $D(r_j,i)$ represents the distance of the $i$-th 3D Gaussian from the rendering plane along the $r_j$. It's important to note that this procedure is designed to simulate the rendering process rather than perform actual rendering.
  

\subsubsection{Reconstructive Generation}
\label{sec:Reconstructive Generation} 
We can use 3D reconstruction techniques to accelerate the creation of realistic surface textures~\cite{kerbl20233d}. We observed that when sampling very small timestep $t$(ranging from $0$ to $200$), the image predicted by Eq.~\ref{Eq:pseudoGT} maintains the same 3D shape as the input image $x_0$
but reveals more detailed and plausible textures. Thus, to maintain shape consistency, we directly \textbf{\textit{generate}} a new 3D representation via \textbf{\textit{3D reconstruction}}~\cite{kerbl20233d}. As depicted in Fig.\ref{fig:Formation Pattern Sampling} (c), after achieving a coarse texture but rich semantic 3D structure, we render 
$K$ images $x_i$, for $i=1,...,K$ from various camera poses $c_i$ around the 3D representation. By adding 
$t$ timestep of noise to these images to obtain $x_{it}$ using Eq.\ref{Eq:ddpm}, we estimate the images
$\hat{x}_{i0}^t$ with plausible textures by Eq.\ref{Eq:pseudoGT}. We then reconstruct them onto the coarse representation by minimizing the following reconstruction loss:
\begin{equation}
L_{rec} = \sum_i ||g(\theta , c_i) - \hat{x}_{i0}^t||_2 .
\end{equation}
This process efficiently generates a representation featuring detailed and plausible textures within \textbf{\textit{15 seconds}}. 
\vspace{-8pt}
\subsection{Camera Sampling}
\label{sec:camera_sampling}
\begin{figure}
\centering
\includegraphics[width=0.46\textwidth]{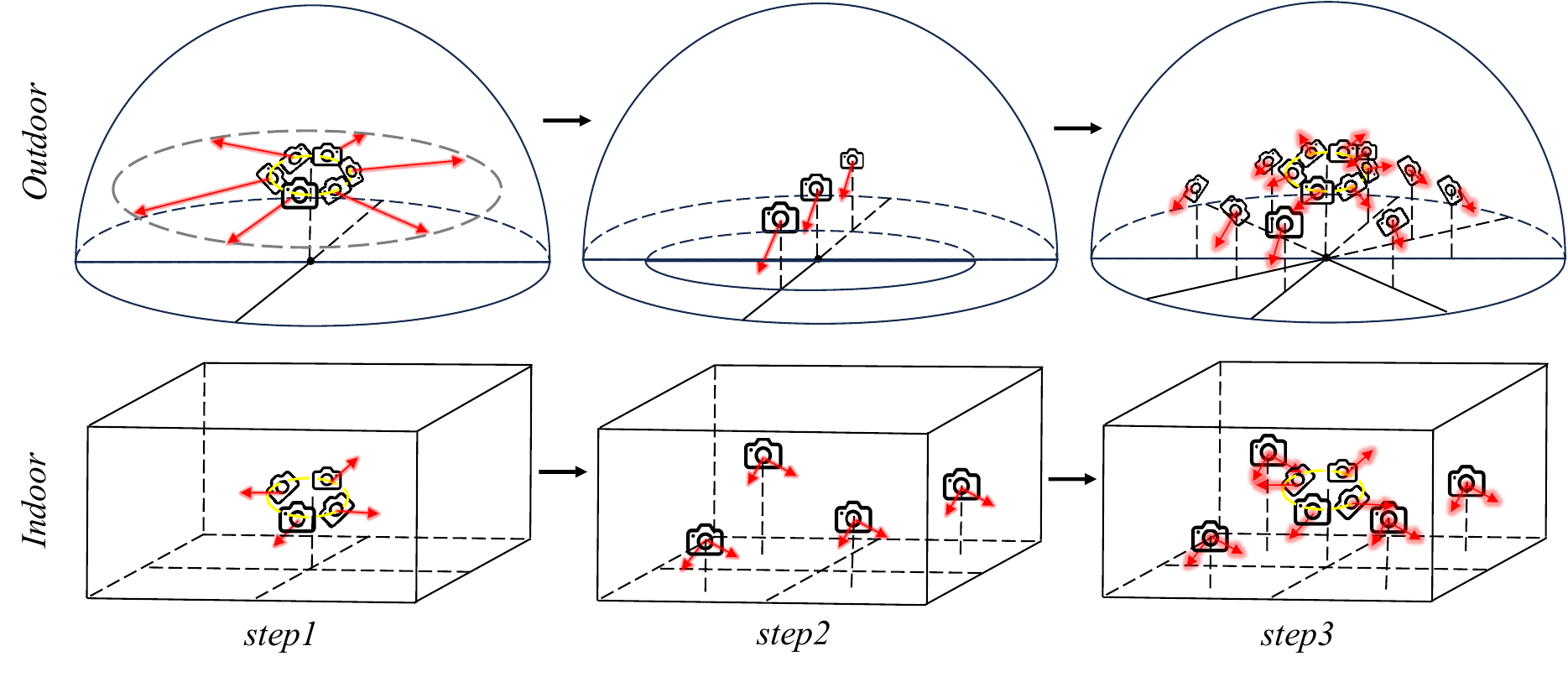}
\centering
\vspace{-8pt}
\caption{Schematic diagram of camera sampling in environment generation.}
\label{fig:camera-sampling}
\end{figure}

\textcolor{black}{Camera sampling is the primary strategy for environmental generation within a scene. Before this, it is necessary to place objects generated in the previous step into the scene based on coordinates derived from Eq.~\ref{eq:coordinate}. This approach prevents the "multi-head" phenomenon in scene generation, where cameras in other methods~\cite{hollein2023text2room,zhang2024text2nerf,ouyang2023text2immersion,wang2024prolificdreamer} cannot truly perceive orientation, resulting in similar content  generated from the same textual prompts at various angles. Consequently, in living rooms generated by some methods~\cite{wang2024prolificdreamer,zhang2024text2nerf,hollein2023text2room,ouyang2023text2immersion}, a sofa appears in every direction as shown in Fig.~\ref{fig:consistency_indoor}. Utilizing the human prior knowledge embedded in GPT-4~\cite{achiam2023gpt}, we have prearranged the layout, ensuring that the rendered scene environment images with information about different objects from different angles, thereby allowing the camera to perceive the room's orientation.}

\par To maintain high quality in scene generation, existing approaches~\cite{hwang2023text2scene,zhang2024text2nerf,ouyang2023text2immersion,cohen2023set,wang2024prolificdreamer} typically restrict camera sampling to a narrow range, which does not provide comprehensive coverage of the scene-wide observations. Employing simple random camera sampling throughout the scene can lead to the breakdown of scene generation during optimization. In response, we have developed a structured, incremental three-step camera sampling strategy, illustrated in Fig.~\ref{fig:camera-sampling}:

\par In the initial stage, we create a basic representation of the  surrounding environment, focusing on indoor walls and distant outdoor elements. \textcolor{black}{We 
lock the parameters of the 3D Gaussians for the ground and objects, limiting camera sampling to coordinates within a certain proximity to the center, to refine the generation of these surroundings.}

\par During the second stage, our focus shifts to generating the coarse ground. At this point, the parameters for the 3D Gaussians representing environments and objects are frozen. For indoor scenes, the space is segmented into distinct regions based on object placement. Camera poses are strategically sampled to target key areas, including objects and the ground, in each iteration. For outdoor scenes, the area is divided into concentric circles determined by their radius. A consistent direction is selected for sampling camera poses around these circles in each iteration, enhancing ground generation. This method ensures thorough coverage of the entire ground area, with a particular focus on zones where the ground meets objects and the surrounding environment.

\par In the third stage, we utilize all previously sampled camera poses to ensure a comprehensive view of the entire scene, focusing on refining all environmental elements. This includes meticulous optimization of parameters for both the ground and the surrounding features. Building on the 3D consistency achieved in earlier two sreps, we then move to the reconstructive generation method in Sec.~\ref{sec:Reconstructive Generation} aimed at acquiring more detailed and plausible textures for the scene.

\par  Camera positions might be obstructed by objects within the scene, requiring collision detection between the camera and these objects. If collisions are detected, the affected camera positions should be discarded to ensure clear visibility.

\subsection{Scene Editing}
\label{sec:scene-editing}
\textcolor{black}{Thanks to  compositional scene generation strategy~\cite{cohen2023set,            vilesov2023cg3d,li20243d}, \sysname\ supports \textbf{flexible} and \textbf{fine-grained} editing of individual objects or environmental elements (e.g., walls, floors, ground), enabling the construction of new scenes through targeted modifications. We organize editing capabilities into three complementary operations: \textit{object relocation}, \textit{appearance modification}, and \textit{temporal movement}.}
\par \noindent \textcolor{black}{\textbf{Object Relocation.} We enable editing by adjusting the object's affine transformation parameters $(s', t', r')$, which control its scale, position, and orientation, respectively. These parameters can be updated without regenerating geometry, allowing fast and lightweight manipulation. Users may provide explicit coordinates or high-level spatial commands (e.g., “move the man backward,” “rotate the chair to face the TV”), which are translated into updated affine parameters. For minor adjustments, such as repositioning a single object, we directly apply the new parameters and verify collision-free placement using  simple AABB collision detection. In cases where multiple objects are significantly repositioned or layout structure is altered, we re-invoke the scene planning module to re-evaluate spatial constraints and update relationships among objects. To maintain scene plausibility, we also sample new camera poses around the relocated objects and re-optimize the local environment (e.g., floor textures or wall geometry) accordingly. This ensures that the resulting scene remains consistent, context-aware, and physically valid after editing. Similarly, when adding a new object, we assign it a valid location using the same constraint-based reasoning. For object removal, we just simply clear its position.}

\textcolor{black}{\textbf{Appearance Modification.} To support high-fidelity object editing, we enable appearance modifications that span both texture geometry refinements. Instead of regenerating the object from scratch, we preserve its existing 3D Gaussians and re-optimize appearance and positional parameters under a new textual description $y_{\text{edit}}$.}
\par \textcolor{black}{We directly adapt the 2D editing process into our MTS method for 3D appearance editing. Traditional 2D editing methods typically consist of two stages: image reconstruction and image editing. In the reconstruction stage, methods such as NTI~\cite{mokady2023null} and PTI~\cite{dong2023prompt} gradually align the latents in the diffusion process to obtain accurate noising and denoising trajectory for the input image. Then, during the editing stage, they inject the target prompt $y_{\text{edit}}$ into the denoising trajectory to guide generation. In our MTS setting, we adopt the same idea on random rendered images in each optimazation. Specifically, we approximate the noising trajectory using DDIM inversion in Eq.~\ref{eq:ddim_inversion} and denosing trajectory using DDIM, just replacing the empty prompt $\emptyset$ with the current object prompt $y$ to obtain an approximate reconstruction trajectory ($x$ represnts the latent in the noising trajectory and $\tilde{x}$ represent the latent in the denoising trajectory):
\begin{equation}
\left\{
\begin{aligned}
    \resizebox{0.89\linewidth}{!}{
        $x_{t_{i+1}} = \sqrt{\bar{\alpha}_{t_{i+1}}} \frac{x_{t_i} - \sqrt{1-\bar{\alpha}_{t_i}}\epsilon_{\theta}(x_{t_i};y,t_i)}{ \sqrt{\bar{\alpha}_{t_i}}} + \sqrt{1-\bar{\alpha}_{t_{i+1}}}\epsilon_{\theta}(x_{t_i};y,t_i)$}
\\
 \resizebox{0.89\linewidth}{!}{
        $\tilde{x}_{t_{i}} = \sqrt{\bar{\alpha}_{t_{i}}} \frac{\tilde{x}_{t_{i+1}} - \sqrt{1-\bar{\alpha}_{t_{i+1}}}\epsilon_{\theta}(\tilde{x}_{t_{i+1}};y,t_{i+1})}{ \sqrt{\bar{\alpha}_{t_{i+1}}}} + \sqrt{1-\bar{\alpha}_{t_{i}}}\epsilon_{\theta}(\tilde{x}_{t_{i+1}};y,t_{i+1})$.}
\end{aligned}
\right.
\end{equation}
Then we directly replace $y$ with $y_{\text{edit}}$ in the denoising process to simulate the 2D editing behavior. This leads to the following MTS-based editing equation:
\begin{equation}
\begin{array}{cc}
 \nabla_{\theta}\mathcal{L}_{\text{MTS\_Editing}}(\theta) = \\ \mathbb{E}_{t,\epsilon,c}  \left[\sum\limits_{i=1}^{m} w(t_i)(\epsilon_{\phi}(x_{t_i}; y_{edit}, t_i)  - 
 \epsilon_{\phi}(x_{t_i}; y, t_i))\frac{\partial g(\theta,c)}{\partial \theta} \right],
\label{eq:MTS_editing}
\end{array}
\end{equation}
, and this can be viewed as guiding the optimization to move away from the original semantics encoded in $y$, and toward those specified by the target prompt $y_{\text{edit}}$.}

\noindent\textcolor{black}{\textbf{Temporal Movement.} To support 4D scene generation with dynamic object motion, we extend the 3D Gaussian representation by introducing a temporal dimension. For static elements such as walls, floors, or backgrounds, the Gaussian parameters remain constant over time. In contrast, for dynamic objects, we apply time-dependent affine transformations $(s_i(t), r_i(t), t_i(t))$ to adjust their position, orientation, and scale at each time step. Given an animation description from the user (e.g., “the man walks from left to right”), a GPT-4 agent automatically generates a \textit{discrete} sequence of affine transformations that simulates a \textit{continuous} trajectory, reflecting the intended motion. This mechanism expands the capability of DreamScene, enabling its application to tasks such as animation creation and virtual environment simulation.}

\vspace{-8pt}
\begin{figure}
\centering
\includegraphics[width=0.48\textwidth]{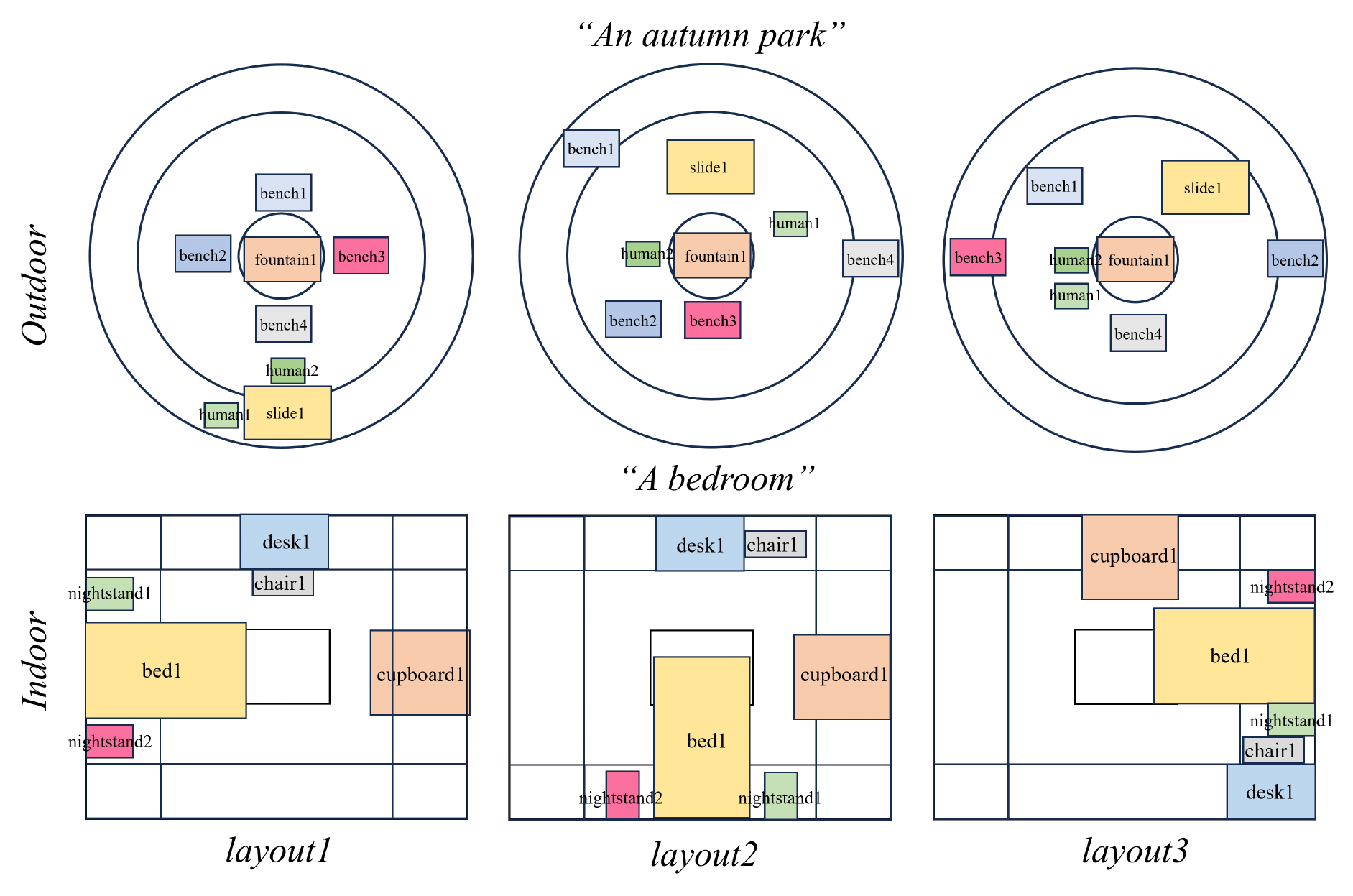}
\centering
\vspace{-5pt}
\caption{Diversity of layout generation.}
\label{fig:layout_result}
\end{figure}

\begin{figure*}
\centering
\includegraphics[width=\textwidth]{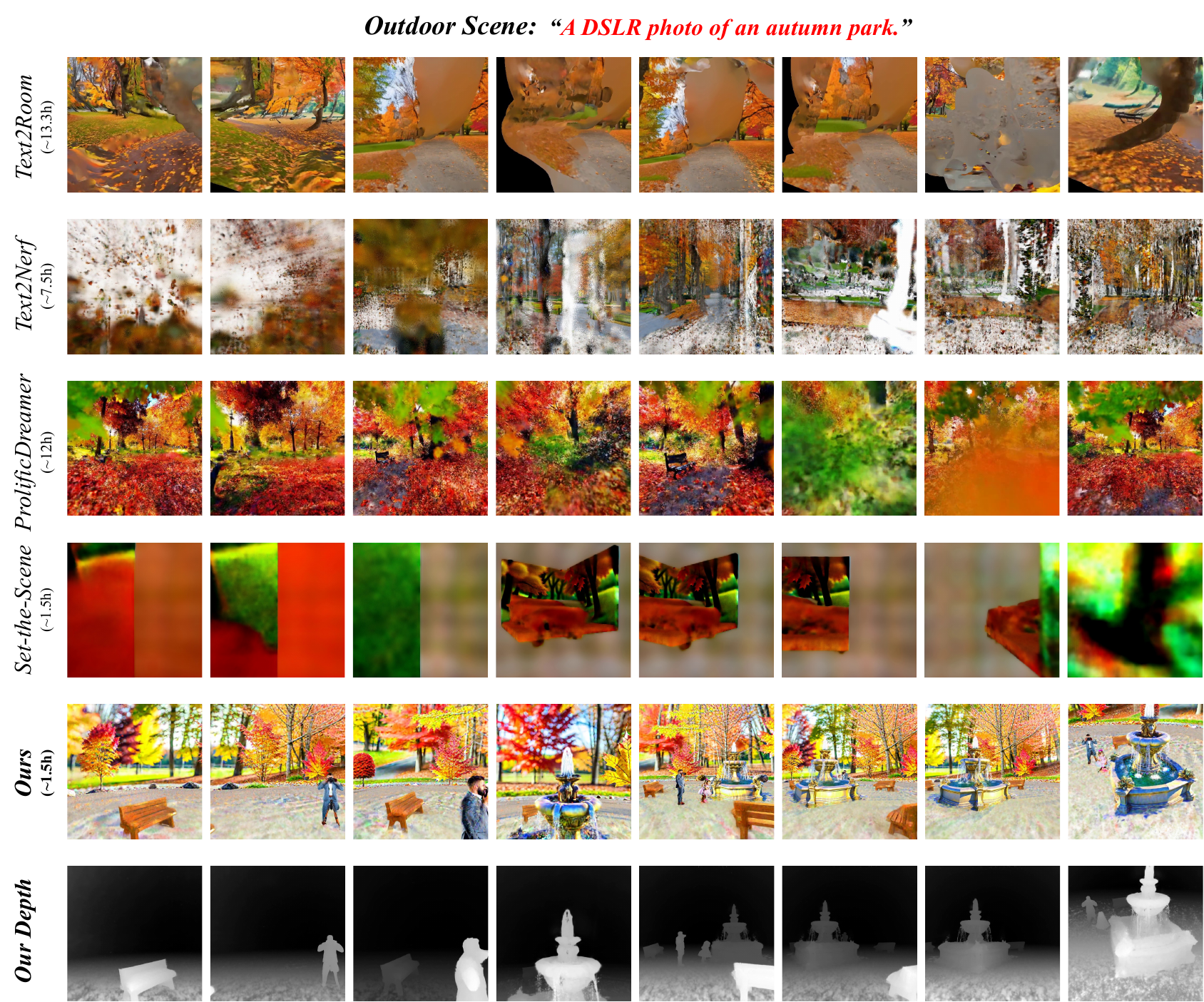}
\centering
\vspace{-8pt}
\caption{\textcolor{black}{Visual consistency and generation quality under diverse scene-wide camera poses in the outdoor scenes.}}
\label{fig:consistency_outdoor}
\end{figure*}

\section{experiment}

\par \noindent \textbf{Implementation Details.} We employ GPT-4~\cite{achiam2023gpt} as our Large Language Model(LLM) for decomposing scene prompts and Point-E~\cite{nichol2022point} for generating initial sparse point clouds of objects. For 2D image generation, we use Stable Diffusion 2.1. The maximum number of iterations for objects is set at 1,500, and for the environment, it is 2,000. The value of the time interval $m$ is $4$. In the reconstructive generation phase, we generate 20 rendering images. To ensure a fair comparison, we tested \sysname\ and all baselines on the same NVIDIA 3090 GPU.
\par \noindent \textbf{Baselines.} For the comparative analysis of text-to-3D scene generation, we utilize the current open-sourced state-of-the-art(SOTA) methods as our baselines: Text2Room~\cite{hollein2023text2room}, Text2NeRF~\cite{zhang2024text2nerf}, ProlificDreamer~\cite{wang2024prolificdreamer}, and Set-the-Scene~\cite{cohen2023set}.
In the domain of text-to-3D generation, our selected baselines are DreamFusion~\cite{poole2022dreamfusion}, Magic3D~\cite{lin2023magic3d}, DreamGaussian~\cite{tang2023dreamgaussian}, and LucidDreamer~\cite{liang2023luciddreamer}(ProlificDreamer, DreamFusion and Magic3D have been reimplemented by Three-studio~\cite{guo2023threestudio}).
\par \noindent \textbf{Evaluation Metrics.} We assessed the generation time for each method~\cite{hollein2023text2room,zhang2024text2nerf,cohen2023set,wang2024prolificdreamer,poole2022dreamfusion,lin2023magic3d,tang2023dreamgaussian,liang2023luciddreamer} and compared the editing capabilities outlined in their respective published papers. \textcolor{black}{We use R-Precision(same setting in DreamTime~\cite{huang2023dreamtime}) to calculate the similarity between the rendered image of the generated 3D representation and the text description.} Additionally, we conducted a user study with 100 participants, where each one rated the quality, consistency, and rationality of the videos on a scale from $1$ to $5$. These 30-second videos were generated by each method across five different scenes—three indoor and two outdoor.
\vspace{-8pt}
\subsection{Qualitative Results}
\label{sec:qualitative results}

\begin{figure*}
\centering
\includegraphics[width=\textwidth]{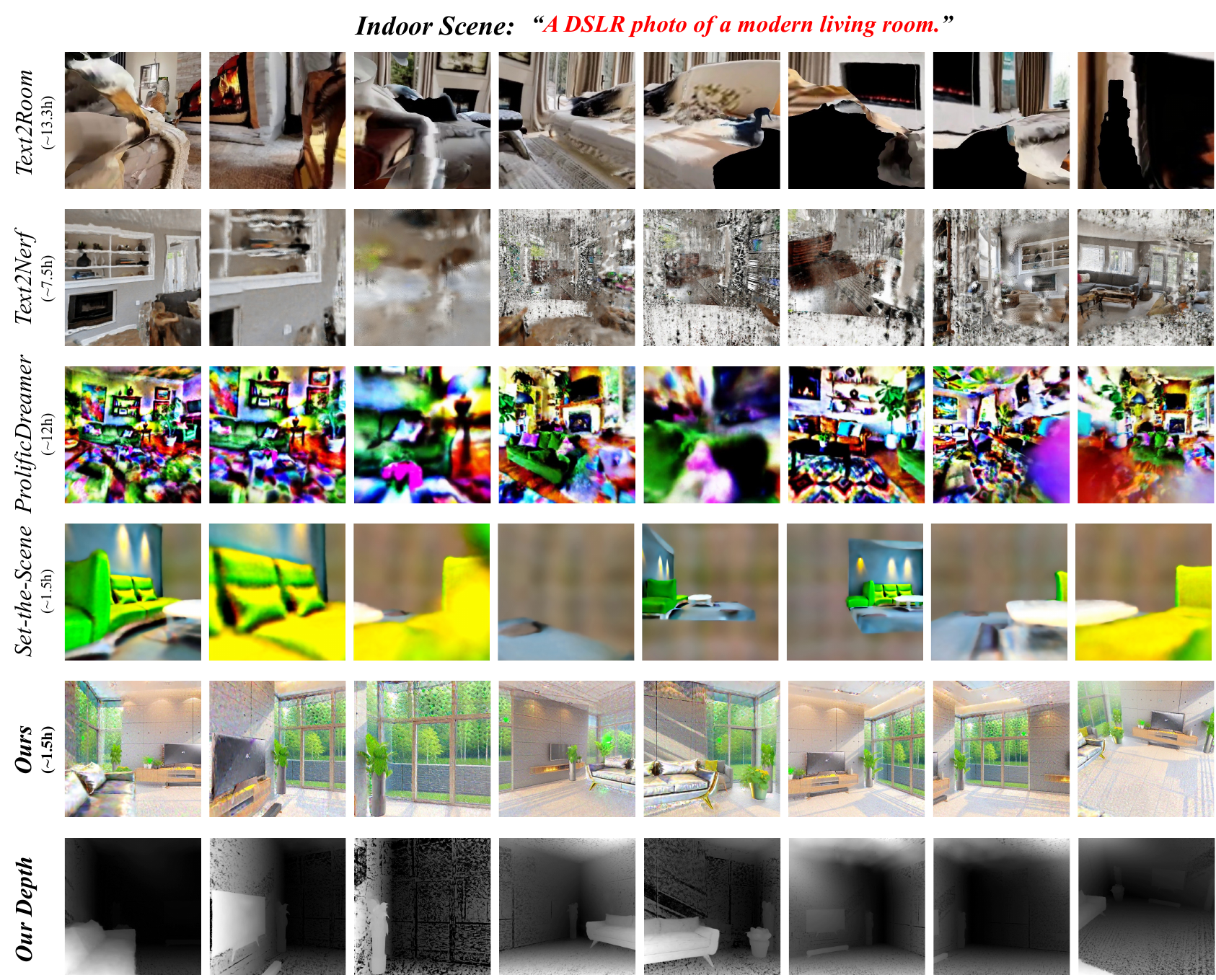}
\centering
\vspace{-8pt}
\caption{\textcolor{black}{Visual consistency and generation quality under diverse scene-wide camera poses in the indoor scenes.}}
\label{fig:consistency_indoor}
\end{figure*}

\begin{figure}
\centering
\includegraphics[width=0.48\textwidth]{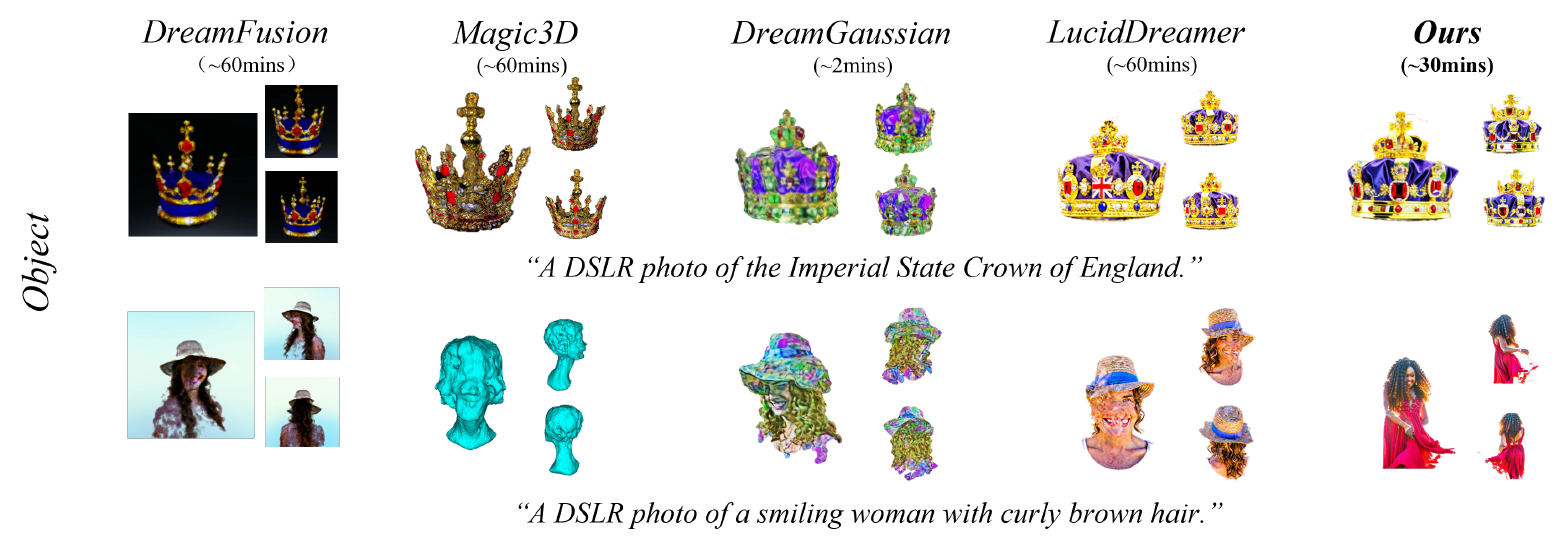}
\vspace{-5pt}
\caption{Comparison with baselines in text-to-3D object generation.}
\label{fig:object_result}
\end{figure}



\begin{figure}
\centering
\includegraphics[width=0.48\textwidth]{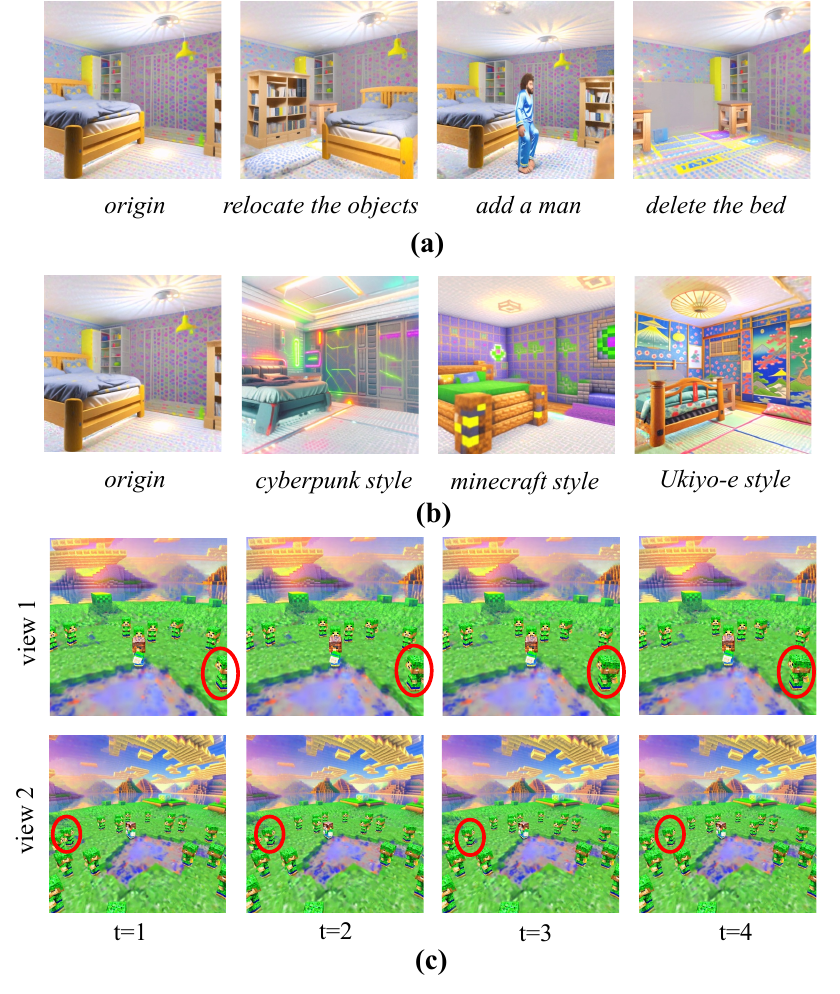}
\centering
\caption{ \textcolor{black}{\sysname\ editing results. (a) shows object-level edits, including relocation, addition, and removal. (b) demonstrates style modifications applied to both objects and environments. (c) presents the \textbf{4D} generation results from multiple viewpoints. }}
\label{fig:editing}
\end{figure}

\noindent \textbf{Layout Generation.} We believe that layout generation should be diverse, as illustrated in Fig.~\ref{fig:layout_result}
, which showcases various layouts for an outdoor park and an indoor bedroom. In the \sysname\ layout generation process, the use of GPT-4's question-and-answer capability results in varied responses each time, although some elements, like the fountain often being at the center of the park, may be consistent. Additionally, during the object placement stage, varying the search order and placement settings (such as centering or edge positioning within an area) contributes to the creation of diverse layouts.

\par \noindent \textbf{Scenes and objects generation.} 
\textcolor{black}{To evaluate scene-wide 3D consistency and generation quality, we conduct comparisons between \sysname\ and representative baselines under diverse camera poses. To ensure fairness, we follow each baseline’s official camera configurations during training. \textit{During testing, we adopt a unified camera trajectory for all methods: the camera first moves in some straight lines across the scene, then circles around the scene center, simulating natural human exploration behavior.} It can be observed that Text2Room~\cite{hollein2023text2room}, Text2NeRF~\cite{zhang2024text2nerf}, and ProlificDreamer~\cite{wang2024prolificdreamer} exhibit poor generalization to novel poses. Even minor viewpoint shifts often lead to severe distortions or structural collapse, indicating a lack of true 3D consistency. In contrast, Set-the-Scene~\cite{cohen2023set}, which shares a similar modular scene composition philosophy with \sysname, achieves relatively stable structure under indoor settings. However, due to its reliance on conventional SDS~\cite{poole2022dreamfusion} optimization, the visual quality is significantly lower and fails to generalize to outdoor scenes. In comparison, as shown in RGB and depth results, \sysname\  generates complete 3D structure, with the best 3D consistency and visual quality among all methods. Additional video results and depth maps from other methods are provided in the supplementary material. }Fig.~\ref{fig:object_result} reveals that our FPS is capable of producing high-quality 3D representations in a brief period, adhering to the text prompts. Although DreamGaussian~\cite{tang2023dreamgaussian} produces results more quickly, it sacrifices the generation quality.

\begin{table}[t]
\caption{quantitative results of \sysname\ compared with baselines. $\uparrow$ means the more the better and $\downarrow$ means the lower the better. q means "quality", c means "consistency" and r means "rationality".}   
\centering

\begin{tabular}{c |c| c |c c  c}  \hline
          Method     & Time &  Editing &\multicolumn{3}{c}{User Study} \\
             &    (hours) $\downarrow$   &   &Q$\uparrow$&C$\uparrow$&R$\uparrow$    \\ \hline
        
        Text2Room~\cite{hollein2023text2room}  & 13.3 & \ding{55} & 2.93 & 2.57 & 2.60 \\ 
        Text2NeRF~\cite{zhang2024text2nerf} &  7.5&  \ding{55} & 3.05 & 2.71 & 2.98 \\ 
         ProlificDreamer~\cite{wang2024prolificdreamer}   & 12.0&\ding{55} & 3.48 & 3.19 & 2.95 \\ 
         Set-the-Scene~\cite{cohen2023set}   &\textbf{1.5}& \checkmark & 2.45 & 3.52 & 2.88 \\ \hline
         \textbf{Ours}   &\textbf{1.5}&   \checkmark & \textbf{3.92} & \textbf{4.24}& \textbf{4.05} \\  \hline
 \end{tabular} 

\label{tab:userstudy}
\end{table}
\begin{table}[t]
 \caption{quantitative results of \sysname\ compared with DreamTime. $\uparrow$ means the more the better.}
\centering
\begin{tabular}{c|c|c|c}
\hline
    R-Precision$\uparrow$ & Ours & Ours(w/o annealing) & 3DGS+DreamTime\\
    \hline
    ViT-L/14 & 71.9\% & 71.9\% & 34\% \\
    ViT-BigG/14 & 70.6\% & 68.6\% & 33.3\% \\
    \hline
    \end{tabular}
\label{tab:dreamtime}
\end{table}
\vspace{-8pt}
\subsection{Quantitative Results}
\label{sec:quantitative-results}
\par \noindent \textbf{Compare with text-to-3D scene methods.} To ensure a fair comparison, we calculate the generation time of our environment generation stage, as the baseline methods~\cite{wang2024prolificdreamer,zhang2024text2nerf,hollein2023text2room} cannot generate objects in the environment independently. The left side of Tab.~\ref{tab:userstudy} demonstrates that our method achieves the shortest generation time for environments with editing capabilities. The right side presents results from a user study, where \sysname\ significantly outperforms the baseline methods~\cite{wang2024prolificdreamer,cohen2023set,zhang2024text2nerf,hollein2023text2room} in terms of consistency and rationality, while maintaining high generation quality.

\par \noindent \textcolor{black}{\textbf{Compare with DreamTime.} 
We use the same evaluation settings as DreamTime~\cite{huang2023dreamtime} to demonstrate that our sampling strategy not only accelerates convergence but also significantly enhances the quality of generation. As illustrated in Tab.~\ref{tab:dreamtime}
, our approach yields better results in terms of CLIP R-Precision after the same 1500 iterations. Additionally, it is observed that the annealing strategy for the time window $T$
 slightly affects the result of generation.}

\begin{figure}
\centering
\includegraphics[width=0.48\textwidth]{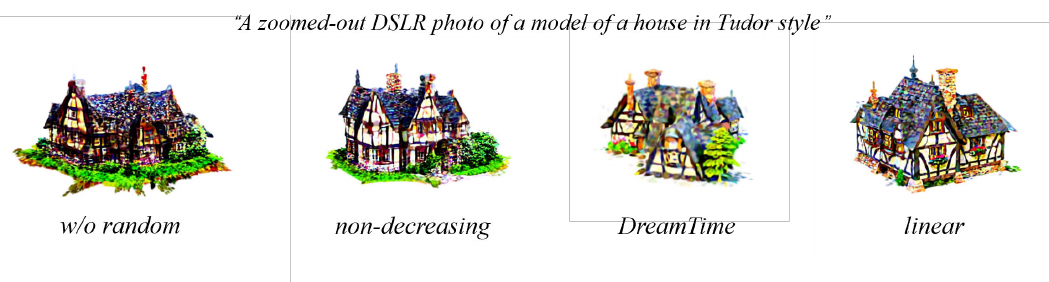}
\vspace{-8pt}
\caption{Ablation results of time window strategy in MTS.}
\label{fig:timewindow}
\end{figure}

\begin{figure}[t]
    \centering
    \includegraphics[width=\linewidth]{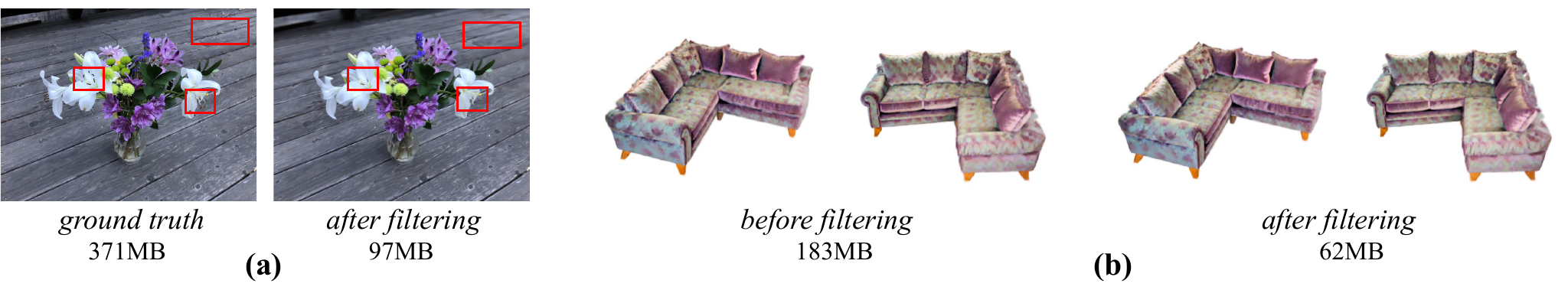}
    \caption{Ablation results of 3D Gaussian filtering algorithm in reconstruction and generation tasks. (a) Data in NeRF-360\cite{mildenhall2021nerf}. (b) Data in generating process.}
 \label{fig:result_faux_rendering}

\end{figure}


\vspace{-8pt}
\subsection{Scene Editing}
\label{sec:scene-editing-result}
\textcolor{black}{Fig.~\ref{fig:editing} showcases the flexible editing capabilities of \sysname, as discussed in Sec.~\ref{sec:scene-editing}. \sysname\ supports object-level relocation, addition, and removal by adjusting the object's affine transformation parameters. During these edits, we resample camera poses at both the original and updated locations to maintain visual consistency. As illustrated in Fig.~\ref{fig:editing}(b), modifying the text prompts enables changes in object appearance or environmental style via Eq.~\ref{eq:MTS_editing}. Furthermore, as shown in Fig.~\ref{fig:editing}(c), by adding temporal control to the affine transformations, we enable continuous object motion over time, achieving \textbf{\textit{4D generation}}. This process also allows multi-view observations of dynamic scenes.}
\subsection{Ablations}
\label{sec:ablations}

\par \noindent \textcolor{black}{ \textbf{Time window strategies in MTS.} As illustrated in Fig.~\ref{fig:timewindow}, the first image demonstrates the result of using fixed-step sampling in MTS rather than random sampling within the time interval. This strategy resulted in notably low quality of generation. Other images depict different strategies for setting time windows in MTS: maintaining a fixed maxstep of $1000$, employing the strategy used in Eq.~\ref{eq:dreamtime}, and using a linearly decreasing strategy. We found that the linearly decreasing strategy outperforms the others. As discussed in Sec.~\ref{sec:Formation Pattern Sampling}, large timesteps $t$ provide valuable semantic information. However, in DreamTime, there are very few sampling points at large $t$. In the later stages of optimization, large $t$ may mislead the optimization direction and result in suboptimal surface outcomes, as seen in the "non-decreasing" strategy.}

\par \noindent \textbf{3D Gaussian filtering.} The method we propose is specifically designed for optimization tasks and can be directly applied to reconstruction tasks~\cite{kerbl20233d} as well. Fig.~\ref{fig:result_faux_rendering} illustrates the outcomes of both reconstruction and generation tasks before and after using the Gaussian filtering algorithm for compression. In the reconstruction task, our method reduced 73.9\% memory consumption for storing 3D Gaussians, at the cost of a slightly blurred image with some loss of detail. Conversely, in the generation task, the compression resulted in a 66.1\% reduction, with no significant loss of quality.
\par
\begin{figure}
    \centering
    \includegraphics[width=\linewidth]{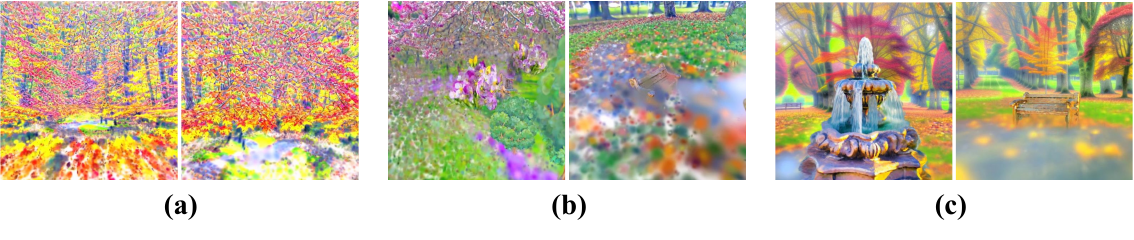}
   \caption{The ablation results of various camera sampling strategies.
 (a) Randomly camera sampling. (b) No distinction between environment and ground. (c) \sysname\ three-step camera sampling strategy}
    \label{fig:sampling_ablation}
\end{figure}

 \noindent \textcolor{black}{\textbf{Camera sampling}. 
Fig.~\ref{fig:sampling_ablation} (a) \textcolor{black}{depicts a scene generated by randomly sampling camera positions within the scene. Due to the challenges in maintaining consistency of scene-wide views at the same location, the optimization process often tends to collapse.} Fig.~\ref{fig:sampling_ablation} (b) adopts a strategy that progresses from the center to the periphery, where the environment and ground are not distinguished. This approach results in improved scene consistency, but the integration between the ground and the scene is poorly executed, and the ground is prone to being populated with coarse 3D Gaussians. Fig.~\ref{fig:sampling_ablation} (c) showcases our three-step strategy, which significantly enhances the quality of generation while ensuring the consistency of both the surrounding environment and the ground.
}



\vspace{-8pt}
\section{Conclusion and Future Work}
\textcolor{black}{We propose \sysname, an end-to-end framework for generating 3D scenes from natural language. The process starts with a scene planning module, where a GPT-4 agent predicts object categories, sizes, descriptions, and spatial relations to build a constraint graph. Based on this, we place objects into the scene with a layout algorithm that ensures reasonable structure and avoids collisions. Then, we generate object geometry using Formation Pattern Sampling, and refine the scene using a three-stage camera sampling strategy for better consistency. \sysname\ also supports scene editing, including moving, adding, or removing objects, changing style, and controlling object motion over time. Our experiments show that \sysname\ can generate consistent, realistic, and editable 3D scenes, making it suitable for a wide range of applications such as VR/AR, Metaverse and simulation.}
\par  \textcolor{black}{In future work, we plan to enhance the scene planning process by capturing more realistic spatial relationships, including fine-grained object placement such as arranging small items on shelves. We also aim to extend the framework to model complex 4D dynamics, including both object-level motion and global scene evolution over time.}

\bibliographystyle{IEEEtran}
\bibliography{IEEEabrv,main}

\begin{thebibliography}{10}
\providecommand{\url}[1]{#1}
\csname url@samestyle\endcsname
\providecommand{\newblock}{\relax}
\providecommand{\bibinfo}[2]{#2}
\providecommand{\BIBentrySTDinterwordspacing}{\spaceskip=0pt\relax}
\providecommand{\BIBentryALTinterwordstretchfactor}{4}
\providecommand{\BIBentryALTinterwordspacing}{\spaceskip=\fontdimen2\font plus
\BIBentryALTinterwordstretchfactor\fontdimen3\font minus \fontdimen4\font\relax}
\providecommand{\BIBforeignlanguage}[2]{{%
\expandafter\ifx\csname l@#1\endcsname\relax
\typeout{** WARNING: IEEEtran.bst: No hyphenation pattern has been}%
\typeout{** loaded for the language `#1'. Using the pattern for}%
\typeout{** the default language instead.}%
\else
\language=\csname l@#1\endcsname
\fi
#2}}
\providecommand{\BIBdecl}{\relax}
\BIBdecl

\bibitem{poole2022dreamfusion}
B.~Poole, A.~Jain, J.~T. Barron, and B.~Mildenhall, ``Dreamfusion: Text-to-3d using 2d diffusion,'' \emph{arXiv preprint arXiv:2209.14988}, 2022.

\bibitem{lin2023magic3d}
C.-H. Lin, J.~Gao, L.~Tang, T.~Takikawa, X.~Zeng, X.~Huang, K.~Kreis, S.~Fidler, M.-Y. Liu, and T.-Y. Lin, ``Magic3d: High-resolution text-to-3d content creation,'' in \emph{Proceedings of the IEEE/CVF Conference on Computer Vision and Pattern Recognition}, 2023, pp. 300--309.

\bibitem{chen2023fantasia3d}
R.~Chen, Y.~Chen, N.~Jiao, and K.~Jia, ``Fantasia3d: Disentangling geometry and appearance for high-quality text-to-3d content creation,'' \emph{arXiv preprint arXiv:2303.13873}, 2023.

\bibitem{liu2023zero}
R.~Liu, R.~Wu, B.~Van~Hoorick, P.~Tokmakov, S.~Zakharov, and C.~Vondrick, ``Zero-1-to-3: Zero-shot one image to 3d object,'' in \emph{Proceedings of the IEEE/CVF International Conference on Computer Vision}, 2023, pp. 9298--9309.

\bibitem{metzer2023latent}
G.~Metzer, E.~Richardson, O.~Patashnik, R.~Giryes, and D.~Cohen-Or, ``Latent-nerf for shape-guided generation of 3d shapes and textures,'' in \emph{Proceedings of the IEEE/CVF Conference on Computer Vision and Pattern Recognition}, 2023, pp. 12\,663--12\,673.

\bibitem{huang2023dreamtime}
Y.~Huang, J.~Wang, Y.~Shi, X.~Qi, Z.-J. Zha, and L.~Zhang, ``Dreamtime: An improved optimization strategy for text-to-3d content creation,'' \emph{arXiv preprint arXiv:2306.12422}, 2023.

\bibitem{yu2023text}
X.~Yu, Y.-C. Guo, Y.~Li, D.~Liang, S.-H. Zhang, and X.~Qi, ``Text-to-3d with classifier score distillation,'' \emph{arXiv preprint arXiv:2310.19415}, 2023.

\bibitem{liang2023luciddreamer}
Y.~Liang, X.~Yang, J.~Lin, H.~Li, X.~Xu, and Y.~Chen, ``Luciddreamer: Towards high-fidelity text-to-3d generation via interval score matching,'' \emph{arXiv preprint arXiv:2311.11284}, 2023.

\bibitem{tang2023dreamgaussian}
J.~Tang, J.~Ren, H.~Zhou, Z.~Liu, and G.~Zeng, ``Dreamgaussian: Generative gaussian splatting for efficient 3d content creation,'' \emph{arXiv preprint arXiv:2309.16653}, 2023.

\bibitem{li2023sweetdreamer}
W.~Li, R.~Chen, X.~Chen, and P.~Tan, ``Sweetdreamer: Aligning geometric priors in 2d diffusion for consistent text-to-3d,'' \emph{arXiv preprint arXiv:2310.02596}, 2023.

\bibitem{nichol2022point}
A.~Nichol, H.~Jun, P.~Dhariwal, P.~Mishkin, and M.~Chen, ``Point-e: A system for generating 3d point clouds from complex prompts,'' \emph{arXiv preprint arXiv:2212.08751}, 2022.

\bibitem{jun2023shap}
H.~Jun and A.~Nichol, ``Shap-e: Generating conditional 3d implicit functions,'' \emph{arXiv preprint arXiv:2305.02463}, 2023.

\bibitem{ramesh2022hierarchical}
A.~Ramesh, P.~Dhariwal, A.~Nichol, C.~Chu, and M.~Chen, ``Hierarchical text-conditional image generation with clip latents,'' \emph{arXiv preprint arXiv:2204.06125}, vol.~1, no.~2, p.~3, 2022.

\bibitem{rombach2022high}
R.~Rombach, A.~Blattmann, D.~Lorenz, P.~Esser, and B.~Ommer, ``High-resolution image synthesis with latent diffusion models,'' in \emph{Proceedings of the IEEE/CVF conference on computer vision and pattern recognition}, 2022, pp. 10\,684--10\,695.

\bibitem{saharia2022photorealistic}
C.~Saharia, W.~Chan, S.~Saxena, L.~Li, J.~Whang, E.~L. Denton, K.~Ghasemipour, R.~Gontijo~Lopes, B.~Karagol~Ayan, T.~Salimans \emph{et~al.}, ``Photorealistic text-to-image diffusion models with deep language understanding,'' \emph{Advances in Neural Information Processing Systems}, vol.~35, pp. 36\,479--36\,494, 2022.

\bibitem{mildenhall2021nerf}
B.~Mildenhall, P.~P. Srinivasan, M.~Tancik, J.~T. Barron, R.~Ramamoorthi, and R.~Ng, ``Nerf: Representing scenes as neural radiance fields for view synthesis,'' \emph{Communications of the ACM}, vol.~65, no.~1, pp. 99--106, 2021.

\bibitem{park2019deepsdf}
J.~J. Park, P.~Florence, J.~Straub, R.~Newcombe, and S.~Lovegrove, ``Deepsdf: Learning continuous signed distance functions for shape representation,'' in \emph{Proceedings of the IEEE/CVF conference on computer vision and pattern recognition}, 2019, pp. 165--174.

\bibitem{kerbl20233d}
B.~Kerbl, G.~Kopanas, T.~Leimk{\"u}hler, and G.~Drettakis, ``3d gaussian splatting for real-time radiance field rendering,'' \emph{ACM Transactions on Graphics}, vol.~42, no.~4, 2023.

\bibitem{muller2022instant}
T.~M{\"u}ller, A.~Evans, C.~Schied, and A.~Keller, ``Instant neural graphics primitives with a multiresolution hash encoding,'' \emph{ACM Transactions on Graphics (ToG)}, vol.~41, no.~4, pp. 1--15, 2022.

\bibitem{shen2021deep}
T.~Shen, J.~Gao, K.~Yin, M.-Y. Liu, and S.~Fidler, ``Deep marching tetrahedra: a hybrid representation for high-resolution 3d shape synthesis,'' \emph{Advances in Neural Information Processing Systems}, vol.~34, pp. 6087--6101, 2021.

\bibitem{cohen2023set}
D.~Cohen-Bar, E.~Richardson, G.~Metzer, R.~Giryes, and D.~Cohen-Or, ``Set-the-scene: Global-local training for generating controllable nerf scenes,'' \emph{arXiv preprint arXiv:2303.13450}, 2023.

\bibitem{hollein2023text2room}
L.~H{\"o}llein, A.~Cao, A.~Owens, J.~Johnson, and M.~Nie{\ss}ner, ``Text2room: Extracting textured 3d meshes from 2d text-to-image models,'' \emph{arXiv preprint arXiv:2303.11989}, 2023.

\bibitem{ouyang2023text2immersion}
H.~Ouyang, K.~Heal, S.~Lombardi, and T.~Sun, ``Text2immersion: Generative immersive scene with 3d gaussians,'' \emph{arXiv preprint arXiv:2312.09242}, 2023.

\bibitem{li2024dreamscene}
H.~Li, H.~Shi, W.~Zhang, W.~Wu, Y.~Liao, L.~Wang, L.-h. Lee, and P.~Zhou, ``Dreamscene: 3d gaussian-based text-to-3d scene generation via formation pattern sampling,'' \emph{arXiv preprint arXiv:2404.03575}, 2024.

\bibitem{zhang2024text2nerf}
J.~Zhang, X.~Li, Z.~Wan, C.~Wang, and J.~Liao, ``Text2nerf: Text-driven 3d scene generation with neural radiance fields,'' \emph{IEEE Transactions on Visualization and Computer Graphics}, 2024.

\bibitem{po2023compositional}
R.~Po and G.~Wetzstein, ``Compositional 3d scene generation using locally conditioned diffusion,'' \emph{arXiv preprint arXiv:2303.12218}, 2023.

\bibitem{wang2024prolificdreamer}
Z.~Wang, C.~Lu, Y.~Wang, F.~Bao, C.~Li, H.~Su, and J.~Zhu, ``Prolificdreamer: High-fidelity and diverse text-to-3d generation with variational score distillation,'' \emph{Advances in Neural Information Processing Systems}, vol.~36, 2024.

\bibitem{zhang2023scenewiz3d}
Q.~Zhang, C.~Wang, A.~Siarohin, P.~Zhuang, Y.~Xu, C.~Yang, D.~Lin, B.~Zhou, S.~Tulyakov, and H.-Y. Lee, ``Scenewiz3d: Towards text-guided 3d scene composition,'' \emph{arXiv preprint arXiv:2312.08885}, 2023.

\bibitem{lin2023componerf}
Y.~Lin, H.~Bai, S.~Li, H.~Lu, X.~Lin, H.~Xiong, and L.~Wang, ``Componerf: Text-guided multi-object compositional nerf with editable 3d scene layout,'' \emph{arXiv preprint arXiv:2303.13843}, 2023.

\bibitem{achiam2023gpt}
J.~Achiam, S.~Adler, S.~Agarwal, L.~Ahmad, I.~Akkaya, F.~L. Aleman, D.~Almeida, J.~Altenschmidt, S.~Altman, S.~Anadkat \emph{et~al.}, ``Gpt-4 technical report,'' \emph{arXiv preprint arXiv:2303.08774}, 2023.

\bibitem{ho2020denoising}
J.~Ho, A.~Jain, and P.~Abbeel, ``Denoising diffusion probabilistic models,'' \emph{Advances in neural information processing systems}, vol.~33, pp. 6840--6851, 2020.

\bibitem{zhou2024gala3d}
X.~Zhou, X.~Ran, Y.~Xiong, J.~He, Z.~Lin, Y.~Wang, D.~Sun, and M.-H. Yang, ``Gala3d: Towards text-to-3d complex scene generation via layout-guided generative gaussian splatting,'' \emph{arXiv preprint arXiv:2402.07207}, 2024.

\bibitem{vilesov2023cg3d}
A.~Vilesov, P.~Chari, and A.~Kadambi, ``Cg3d: Compositional generation for text-to-3d via gaussian splatting,'' \emph{arXiv preprint arXiv:2311.17907}, 2023.

\bibitem{lan20242d}
K.~Lan, H.~Li, H.~Shi, W.~Wu, L.~Wang, and Y.~Liao, ``2d-guided 3d gaussian segmentation,'' in \emph{2024 Asian Conference on Communication and Networks (ASIANComNet)}.\hskip 1em plus 0.5em minus 0.4em\relax IEEE, 2024, pp. 1--5.

\bibitem{barron2021mip}
J.~T. Barron, B.~Mildenhall, M.~Tancik, P.~Hedman, R.~Martin-Brualla, and P.~P. Srinivasan, ``Mip-nerf: A multiscale representation for anti-aliasing neural radiance fields,'' in \emph{Proceedings of the IEEE/CVF International Conference on Computer Vision}, 2021, pp. 5855--5864.

\bibitem{shi2023mvdream}
Y.~Shi, P.~Wang, J.~Ye, M.~Long, K.~Li, and X.~Yang, ``Mvdream: Multi-view diffusion for 3d generation,'' \emph{arXiv preprint arXiv:2308.16512}, 2023.

\bibitem{li2024text}
H.~Li, Y.~Tian, Y.~Wang, Y.~Liao, L.~Wang, Y.~Wang, and P.~Y. Zhou, ``Text-to-3d generation by 2d editing,'' \emph{arXiv preprint arXiv:2412.05929}, 2024.

\bibitem{yi2023gaussiandreamer}
T.~Yi, J.~Fang, G.~Wu, L.~Xie, X.~Zhang, W.~Liu, Q.~Tian, and X.~Wang, ``Gaussiandreamer: Fast generation from text to 3d gaussian splatting with point cloud priors,'' \emph{arXiv preprint arXiv:2310.08529}, 2023.

\bibitem{mokady2023null}
R.~Mokady, A.~Hertz, K.~Aberman, Y.~Pritch, and D.~Cohen-Or, ``Null-text inversion for editing real images using guided diffusion models,'' in \emph{Proceedings of the IEEE/CVF Conference on Computer Vision and Pattern Recognition}, 2023, pp. 6038--6047.

\bibitem{hertz2022prompt}
A.~Hertz, R.~Mokady, J.~Tenenbaum, K.~Aberman, Y.~Pritch, and D.~Cohen-Or, ``Prompt-to-prompt image editing with cross attention control,'' \emph{arXiv preprint arXiv:2208.01626}, 2022.

\bibitem{zhou2024layout}
J.~Zhou, X.~Li, L.~Qi, and M.-H. Yang, ``Layout-your-3d: Controllable and precise 3d generation with 2d blueprint,'' \emph{arXiv preprint arXiv:2410.15391}, 2024.

\bibitem{nath2025decompdreamer}
U.~Nath, R.~Goel, R.~Khurana, K.~Min, M.~Ollila, P.~Turaga, V.~Jampani, and T.~Gowda, ``Decompdreamer: Advancing structured 3d asset generation with multi-object decomposition and gaussian splatting,'' \emph{arXiv preprint arXiv:2503.11981}, 2025.

\bibitem{bahmani2023cc3d}
S.~Bahmani, J.~J. Park, D.~Paschalidou, X.~Yan, G.~Wetzstein, L.~Guibas, and A.~Tagliasacchi, ``Cc3d: Layout-conditioned generation of compositional 3d scenes,'' in \emph{Proceedings of the IEEE/CVF International Conference on Computer Vision}, 2023, pp. 7171--7181.

\bibitem{zhang2024berfscene}
Q.~Zhang, Y.~Xu, Y.~Shen, B.~Dai, B.~Zhou, and C.~Yang, ``Berfscene: Bev-conditioned equivariant radiance fields for infinite 3d scene generation,'' in \emph{Proceedings of the IEEE/CVF Conference on Computer Vision and Pattern Recognition}, 2024, pp. 6839--6849.

\bibitem{paschalidou2021atiss}
D.~Paschalidou, A.~Kar, M.~Shugrina, K.~Kreis, A.~Geiger, and S.~Fidler, ``Atiss: Autoregressive transformers for indoor scene synthesis,'' \emph{Advances in Neural Information Processing Systems}, vol.~34, pp. 12\,013--12\,026, 2021.

\bibitem{fu2024scene}
R.~Fu, J.~Liu, X.~Chen, Y.~Nie, and W.~Xiong, ``Scene-llm: Extending language model for 3d visual understanding and reasoning,'' \emph{arXiv preprint arXiv:2403.11401}, 2024.

\bibitem{hong20233d}
Y.~Hong, H.~Zhen, P.~Chen, S.~Zheng, Y.~Du, Z.~Chen, and C.~Gan, ``3d-llm: Injecting the 3d world into large language models,'' \emph{Advances in Neural Information Processing Systems}, vol.~36, pp. 20\,482--20\,494, 2023.

\bibitem{wang2024root}
Y.~Wang, S.-Y. Chen, Z.~Zhou, S.~Li, H.~Li, W.~Zhou, and H.~Li, ``Root: Vlm based system for indoor scene understanding and beyond,'' \emph{arXiv preprint arXiv:2411.15714}, 2024.

\bibitem{song2020denoising}
J.~Song, C.~Meng, and S.~Ermon, ``Denoising diffusion implicit models,'' \emph{arXiv preprint arXiv:2010.02502}, 2020.

\bibitem{ho2022classifier}
J.~Ho and T.~Salimans, ``Classifier-free diffusion guidance,'' \emph{arXiv preprint arXiv:2207.12598}, 2022.

\bibitem{chen2024survey}
G.~Chen and W.~Wang, ``A survey on 3d gaussian splatting,'' \emph{arXiv preprint arXiv:2401.03890}, 2024.

\bibitem{wu2024consistent3d}
Z.~Wu, P.~Zhou, X.~Yi, X.~Yuan, and H.~Zhang, ``Consistent3d: Towards consistent high-fidelity text-to-3d generation with deterministic sampling prior,'' in \emph{Proceedings of the IEEE/CVF Conference on Computer Vision and Pattern Recognition}, 2024, pp. 9892--9902.

\bibitem{dong2023prompt}
W.~Dong, S.~Xue, X.~Duan, and S.~Han, ``Prompt tuning inversion for text-driven image editing using diffusion models,'' in \emph{Proceedings of the IEEE/CVF International Conference on Computer Vision}, 2023, pp. 7430--7440.

\bibitem{fan2023lightgaussian}
Z.~Fan, K.~Wang, K.~Wen, Z.~Zhu, D.~Xu, and Z.~Wang, ``Lightgaussian: Unbounded 3d gaussian compression with 15x reduction and 200+ fps,'' \emph{arXiv preprint arXiv:2311.17245}, 2023.

\bibitem{lee2023compact}
J.~C. Lee, D.~Rho, X.~Sun, J.~H. Ko, and E.~Park, ``Compact 3d gaussian representation for radiance field,'' \emph{arXiv preprint arXiv:2311.13681}, 2023.

\bibitem{hwang2023text2scene}
I.~Hwang, H.~Kim, and Y.~M. Kim, ``Text2scene: Text-driven indoor scene stylization with part-aware details,'' in \emph{Proceedings of the IEEE/CVF Conference on Computer Vision and Pattern Recognition}, 2023, pp. 1890--1899.

\bibitem{li20243d}
H.~Li, L.~Ma, H.~Shi, Y.~Hao, Y.~Liao, L.~Cheng, and P.~Y. Zhou, ``3d-goi: 3d gan omni-inversion for multifaceted and multi-object editing,'' in \emph{European Conference on Computer Vision}.\hskip 1em plus 0.5em minus 0.4em\relax Springer, 2024, pp. 390--406.

\bibitem{guo2023threestudio}
Y.-C. Guo, Y.-T. Liu, C.~Wang, Z.-X. Zou, G.~Luo, C.-H. Chen, Y.-P. Cao, and S.-H. Zhang, ``threestudio: A unified framework for 3d content generation,'' 2023.

\end{thebibliography}
\appendices

\section{Theoretical Derivation of Multi-timestep Sampling (MTS)}
Our Multi-timestep Sampling (MTS) strategy is grounded in a key empirical observation in diffusion-based generation: different timesteps encode information at varying levels of semantic granularity. This motivates the use of multiple denoising steps to improve generation quality and optimization stability. In this section, we present a theoretical analysis of MTS and establish its connection to diffusion-based 2D editing methods. This analysis also confirms that MTS is not a heuristic mechanism, but a principled strategy supported by the underlying behavior of diffusion models.
\subsection*{1.Derivation and Approximation}
\label{sec:Approximation}
We first obtain a latent noisy trajectory $x_{t_0},x_{t_1},...,x_{t_m}$ using DDIM Inversion as follows:
\begin{equation}
\begin{aligned}
       x_{t_{i+1}} = \sqrt{\bar{\alpha}_{t_{i+1}}} \frac{x_{t_i} - \sqrt{1-\bar{\alpha}_{t_i}}\epsilon_{\theta}(x_{t_i},t_i,\emptyset)}{ \sqrt{\bar{\alpha}_{t_i}}} \\ +\sqrt{1-\bar{\alpha}_{t_{i+1}}}\epsilon_{\theta}(x_{t_i},t_i,\emptyset),
\end{aligned}
\label{eq:ddim_inversion_2}
\end{equation}
where $\emptyset$ is an empty prompt used to preserve the original image content.

We then denoise the latents along the trajectory using DDIM:
\begin{equation}
\begin{aligned}
       \tilde{x}_{t_{i}} = \sqrt{\bar{\alpha}_{t_{i}}} \frac{x_{t_{i+1}} - \sqrt{1-\bar{\alpha}_{t_{i+1}}}\tilde{\epsilon}_{\theta}(x_{t_{i+1}},t_{i+1},y,\emptyset)}{ \sqrt{\bar{\alpha}_{t_{i+1}}}} \\+ \sqrt{1-\bar{\alpha}_{t_{i}}}\tilde{\epsilon}_{\theta}(x_{t_{i+1}},t_{i+1},y,\emptyset),
\end{aligned}
\label{eq:ddim}
\end{equation}
\begin{equation}
\tilde{\epsilon}_{\theta}(x_t,t,\emptyset,y) =  \epsilon_{\theta}(x_t,t,\emptyset) + \lambda(\epsilon_{\theta}(x_t,t,y) - \epsilon_{\theta}(x_t,t,\emptyset)),
\label{eq:CFG}
\end{equation}
where $y$ is the target prompt and $\lambda$ is the guidance scale.

By simplifying Eq.~\ref{eq:ddim_inversion_2} and Eq.~\ref{eq:ddim}, we obtain:
\begin{equation}
\resizebox{0.9\linewidth}{!}{
$
\left\{
\begin{aligned}
     \frac{x_{t_{i+1}}}{\sqrt{\overline{\alpha}_{t_{i+1}}}} - \frac{x_{t_{i}}}{\sqrt{\overline{\alpha}_{t_{i}}}} &= (\sqrt{\frac{1-\overline{\alpha}_{t_{i+1}}}{\overline{\alpha}_{t_{i+1}}}}-\sqrt{\frac{1-\overline{\alpha}_{t_{i}}}{\overline{\alpha}_{t_{i}}}})\epsilon_\theta(x_{t_{i}},t_{i},\emptyset), \\
    \frac{x_{t_{i+1}}}{\sqrt{\overline{\alpha}_{t_{i+1}}}} - \frac{\tilde{x}_{t_{i}}}{\sqrt{\overline{\alpha}_{t_{i}}}} &= (\sqrt{\frac{1-\overline{\alpha}_{t_{i+1}}}{\overline{\alpha}_{t_{i+1}}}}-\sqrt{\frac{1-\overline{\alpha}_{t_{i}}}{\overline{\alpha}_{t_{i}}}})\tilde{\epsilon}_\theta(x_{t_{i+1}},t_{i+1},y,\emptyset),
\end{aligned}
\right.
$}
\end{equation}
Subtracting the two equations gives:
\begin{equation}
\begin{aligned}
    x_{t_i} - \tilde{x}_{t_i} = (\sqrt{\frac{1-\overline{\alpha}_{t_{i+1}}}{\overline{\alpha}_{t_{i+1}}}} - \sqrt{\frac{1-\overline{\alpha}_{t_{i}}}{\overline{\alpha}_{t_{i}}}}) \\ \times(\tilde{\epsilon}_\theta(x_{t_{i+1}},t_{i+1},y,\emptyset) - \epsilon_\theta(x_{t_{i}},t_{i},\emptyset)).
\end{aligned}
\label{eq:t_i-t_i}
\end{equation}

When $t_{i+1}$ is close to $t_i$, we can approximate:
\begin{equation}
\begin{aligned}
    \tilde{\epsilon}_\theta(x_{t_{i+1}},t_{i+1},y,\emptyset) \approx \tilde{\epsilon}_\theta(x_{t_{i}},t_{i},y,\emptyset).
\end{aligned}
\label{eq:approximation}
\end{equation}

Substituting Eq.~\ref{eq:approximation} into Eq.~\ref{eq:t_i-t_i} yields:
\begin{equation}
\resizebox{0.9\linewidth}{!}{$
\begin{aligned}
  x_{t_i} - \tilde{x}_{t_i} & \approx (\sqrt{\frac{1-\overline{\alpha}_{t_{i+1}}}{\overline{\alpha}_{t_{i+1}}}}-\sqrt{\frac{1-\overline{\alpha}_{t_{i}}}{\overline{\alpha}_{t_{i}}}})(\tilde{\epsilon}_\theta(x_{t_{i}},t_{i},y,\emptyset)-\epsilon_\theta(x_{t_{i}},t_{i},\emptyset)) \\
  &=(\sqrt{\frac{1-\overline{\alpha}_{t_{i+1}}}{\overline{\alpha}_{t_{i+1}}}}-\sqrt{\frac{1-\overline{\alpha}_{t_{i}}}{\overline{\alpha}_{t_{i}}}})\\&
  \times(\epsilon_{\theta}(x_{t_i},t_i,\emptyset) + \lambda(\epsilon_{\theta}(x_{t_i},t_i,y) - \epsilon_{\theta}(x_{t_i},t_i,\emptyset))-\epsilon_\theta(x_{t_{i}},t_{i},\emptyset))\\
  &=\lambda(\sqrt{\frac{1-\overline{\alpha}_{t_{i+1}}}{\overline{\alpha}_{t_{i+1}}}}-\sqrt{\frac{1-\overline{\alpha}_{t_{i}}}{\overline{\alpha}_{t_{i}}}})\\&
  \times(\epsilon_{\theta}(x_{t_i},t_i,y) - \epsilon_{\theta}(x_{t_i},t_i,\emptyset))
  \label{eq:approx}
\end{aligned}
$}
\end{equation}

Thus, we have:
\[
x_{t_i} - \tilde{x}_{t_i} \propto \epsilon_{\theta}(x_{t_i},t_i,y) - \epsilon_{\theta}(x_{t_i},t_i,\emptyset),
\]
Therefore, it can be regarded as $x_{t_i} - \tilde{x}_{t_i} \propto (\epsilon_{\theta}(x_{t_i},t_i,y) - \epsilon_{\theta}(x_{t_i},t_i,\emptyset)) $, where $(\epsilon_{\theta}(x_{t_i},t_i,y) - \epsilon_{\theta}(x_{t_i},t_i,\emptyset))$ is the information at timestep $t_i$ in MTS.   
\par In fact, this approximation in Eq.~\ref{eq:approximation} introduces certain errors, which become more significant as the  $\Delta T=t_{i+1}-t_i$
  increases, as illustrated in Fig.~\ref{fig:stepsize}. Therefore, reducing $\Delta T$ leads to higher generation quality. However, this also increases the number of diffusion steps, resulting in higher computational cost. Considering computational constraints, we set $\Delta T$  to $50\sim100$ in our implementation.
\subsection*{2. Connection to 2D Editing}
Next, we interpret $x_{t_i} - \tilde{x}_{t_i}$ from the perspective of 2D image editing using diffusion models.
\par Text-guided 2D image editing aims to modify an input image according to a target text prompt. Existing diffusion-based 2D editing methods generally consist of two main stages. The first stage is inversion, which focuses on preserving the content of the input image. This is typically done by aligning a complete noising and denoising trajectory, enabling faithful reconstruction of the original image. During the noising process, DDIM Inversion with an empty text prompt is often used to preserve the input image’s content. The denoising path is then aligned through optimization over text embeddings at each timestep. The second stage is editing, which aims to inject the semantic content of the target text into the input image. In this stage, the image is progressively denoised using the target text prompt, which naturally integrates new content into the reconstructed image. Multi-step trajectory modeling is also critical in 2D editing. In the inversion stage, it helps align content across multiple granularities to enhance reconstruction. In the editing stage, injecting the target text across timesteps allows fine-grained control over the strength and scope of the edits~\cite{hertz2022prompt}.

Under a similar setting to MTS, these ,method denote the noising trajectory as $\tilde{x}_{t_0},\tilde{x}_{t_1},...,\tilde{x}_{t_m}$ and the denoising trajectory as $\tilde{x}_{t_0},\tilde{x}_{t_1},...,\tilde{x}_{t_m}$ In the inversion stage, the goal is to align these two trajectories to reconstruct the original image. Since the exact prompt that describes the input image is unknown, recent approaches (e.g., NTI~\cite{mokady2023null}, PTI~\cite{dong2023prompt}) leverage differentiable null-text prompts $ \emptyset_t$ or conditional target texts $y_t$
  to optimize this alignment. This process can be formulated as:
\begin{equation}
   \alpha_{t_i} = \arg \min\limits_{ \alpha_{t_i}} ||x_{t_{i}} - \tilde{x}_{t_{i}}(t_i, \alpha_{t_i})||_2^2,
\end{equation}
where $i=m,...,0$ and $\alpha_{t_i} = \emptyset_{t_i}$ or $y_{t_i}$. This alignment process is typically achieved by minimizing the difference $x_{t_{i}} - \tilde{x}_{t_{i}}$,  effectively guiding $ \tilde{x}_{t_{i}} \rightarrow x_{t_{i}}$. In MTS, we observe a similar mechanism. As shown in Eq.\ref{eq:approx}, the difference $\epsilon_{\theta}(x_{t_i},t_i,y) - \epsilon_{\theta}(x_{t_i},t_i,\emptyset)$ is proportional to $x_{t_{i}} - \tilde{x}_{t_{i}}$ except that the direction is reversed: we aim to move $x_{t_{i}} \rightarrow \tilde{x}_{t_{i}}$, since $\tilde{x}_{t_{i}}$ contains semantic information from the target text prompt and this information needs to be backpropagated through 
$x_{t_{i}}$ into the 3D representation.

Editing methods align with multi-step denoising trajectories in diffusion processes to produce high-quality images. This alignment mechanism similarly enables MTS to align with high-quality denoising trajectories, thereby achieving efficient generation. It also explains why traditional SDS~\cite{poole2022dreamfusion} methods tend to produce oversaturated results: they typically use a single-step denoising process with a large timestep , which leads to coarse and imprecise supervision. In contrast, standard diffusion models perform multi-step denoising with smaller timestep, allowing for more accurate approximation of the underlying data distribution.

\section{Scene Planning Template}
We use the prompts shown in Fig.~\ref{fig:prompt-1}, Fig.~\ref{fig:prompt-2}, and Fig.~\ref{fig:prompt-3} to obtain structured information from GPT-4~\cite{achiam2023gpt}, which is then parsed using Python. From the user's open-ended prompt or dialogue, we extract the corresponding \textcolor{red}{\{User Constraint\}}. We prepend each prompt with the instruction: "You are a professional scene designer. Based on the user requirements \textcolor{red}{{User Constraint}}, and your domain knowledge..." This approach allows us to leverage both the user’s specific intent and GPT-4’s rich scene prior knowledge.

\section{Scene Planning Template}
We provide a detailed algorithmic description of the training process of DreamScene as shown in ~\ref{alg:DreamScene}. 

\begin{figure*}[htbp]
\begin{mybox}{Object Information Prompt}
\footnotesize
You are a professional scene designer. Based on the user requirements \textcolor{red}{{User Constraint}} and your domain knowledge, your task is to generate a list of objects commonly found in the described scene.  For each object, please include its frequency of appearance, typical dimensions ([x, y, z] in meters), and a brief description starting with "A DSLR photo of". Ensure that the object descriptions are consistent with the scene's style and reflect common human understanding. Output should be formatted as follows in JSON:\\ 
\textbf{Input}: \\
a living room \\
\textbf{Output}: \\
\{"sofa": \{"number":2, "size":[2.0,1.0,0.8], "description":"A DSLR photo of a plush, grey sectional sofa, featuring deep cushions and soft fabric."\}, \\"coffee table":\{"number":1, "size":[1.5,1.0,0.5], "description": "A DSLR photo of a round, glass-top coffee table with a modern design and a sturdy metal base."\}, \\
"TV":\{"number":1, "size":[1.4, 0.8, 0.1], "description": "A DSLR photo of a large flat-screen TV, featuring a wide, slim display on the TV stand."\}, \\
"TV stand": \{"number":1, "size":[1.0, 0.4, 0.5], "description": "A DSLR photo of a sleek, modern TV stand featuring open shelving and a minimalist design."\} \\
"potted plant": \{"number":2, "size":[0.5, 0.5, 1.0], "description": "A DSLR photo of a vibrant, lush plant with broad green leaves in a decorative pot."\} \}
\\ Now, let's design the scene: \{input\}.
\end{mybox}
\vspace{-5pt}
\caption{Prompt template for object information with GPT-4.
}
\label{fig:prompt-1}
\end{figure*}

\begin{figure*}[htbp]
\begin{mybox}{Layout Information Prompt}
\footnotesize
You are a scene placement expert. Based on the user requirements \textcolor{red}{{User Constraint}} and your domain knowledge, your task is to determine the spatial relationship between an object and its environment based on the object’s name and common human understanding. There are four relationships to choose from: 1. CENTER, the object is in the center of the scene 2. SIDE, the object is at the boundary of the scene 3. CORNER, the object is in the corner of the scene 4. OTHERS, the object is in other places. When dealing with multiple similar objects, arrange their positions reasonably to prevent conflicts. Please return in the following example format in JSON format.\\ 
\textbf{Input}: \\
{"scene\_type":"indoor scene", "scene\_text":"a living room", "objects\_list":["sofa1", 
 "sofa2", "coffee table1", "TV1","TV stand1", "potted plant1", "potted plant2"]}\\
\textbf{Output}: \\
\{"sofa1": SIDE, "sofa2": SIDE, "coffee table1": CENTER, "TV1": SIDE, "TV stand1": SIDE, "potted plant1": CORNER, "potted plant2": CORNER\}\\
Now, I need select for \{input\}.
\end{mybox}
\vspace{-5pt}
\caption{Prompt template for layout information with GPT-4.
}
\label{fig:prompt-2}
\end{figure*}

\begin{figure*}[htbp]
\begin{mybox}{Objects Constraints Prompt}
\footnotesize
You are an expert in scene arrangement. Based on the user requirements \textcolor{red}{{User Constraint}}, the given environment, and your domain knowledge, your task is to select objects from the provided list that are relevant to the current object based on common human usage, and describe their spatial or functional relationships. The possible relationships include: 1.LEFT, indicating the current object is at the left of the selected object. 1.RIGHT, indicating the current object is at the right of the selected object. 3.FRONT, indicating the current object is at the front of the selected object. 4.BEHIND, indicating the current object is at the behind of the selected object. 5.OVER, indicating the current object is above the selected object. 6.UNDER, indicating the current object is below the selected object. 7.NEXT, indicating the current object is near the selected object. 8.OPPOSITE, indicating the current object is opposite the selected object. Output the selected object and their relationship in JSON format. For example:\\ 
\textbf{Input}: \\
{"scene\_type": "indoor scene", "scene\_text": "a living room","current\_object": "sofa1", "objects\_list": ["sofa2","coffee table1","TV1" "TV stand1", "potted plant1", "potted plant2"]}\\
\textbf{Output}: \\
\{"sofa2": NEXT, "coffee table1": FRONT, "TV1": OPPOSITE, "TV stand1": OPPOSITE\}\\
Now, I need design for \{input\}.
\end{mybox}
\vspace{-5pt}
\caption{Prompt template for objects constraints with GPT-4.
}
\label{fig:prompt-3}
\end{figure*}

\begin{algorithm}
\caption{\sysname}
\label{alg:DreamScene}
\begin{algorithmic}[1] 
\small
\State Input: A simple scene text $y_S$, the maximum number of iteration $iter_m$, iteration for Gaussian filtering $iter_f$, the number of intervals $m$, compression ratio $\eta$, $x$ are the coordinates for 3D Gaussians~\cite{kerbl20233d}. 
\State Initialize Stable Diffusion~\cite{metzer2023latent}, Point-E~\cite{nichol2022point};
\State Generate objects descriptions $y_1,y_2,...,y_N$ and layouts $l_1,l_2,...,l_N$($l=[s, t, r]$, $s$ is the scale coefficient, $t$ is the  translation coefficient and $r$ is the roation coefficient) by Scene Planning Module;
\For{$n$ in $[1,2,...,N, S]$}
    \If{$n$ is not $S$}
        \State Initialize 3D Gaussian of $obj_n$ by Point-E 
    \Else
        \State Initialize cuboid or hemispherical 3D Gaussian for the  scene
    \EndIf
    \For{$iter = [0,1,...,max\_iter]$}
        \If{$n$ is not $S$}
            \State Spherical sample camera pose $c$ 
        \Else
            \State Sample camera pose $c$ following strategy in Sec.IV-C
        \EndIf
        \State $x_0 = g(\theta,c)$
        \State $T_{end} = (1-\frac{iter}{iter_m}) \times 1000$
        \For{$i = [1,2,...,m]$}
            \State $t_i = T_{end}\cdot random(\frac{i-1}{m},\frac{i}{m})$
            \State $x_i=DDIM_(x_{i-1},i)$
            \State $\epsilon_{\phi}(x_{t_i}; y_n, t_i) = $U-Net$(x_{t_i},y_n,t_i)$
            \State $\epsilon_{\phi}(x_{t_i}; \emptyset, t_i) = $U-Net$(x_{t_i},\emptyset,t_i)$ 
        \EndFor
        \State $\nabla_{\theta}\mathcal{L}_{\text{MTS}}(\theta) = $
        \State $\mathbb{E}_{t,\epsilon,c}\left[\sum\limits_{i=1}^{m} w(t_i)(\epsilon_{\phi}(x_{t_i}; y_n, t_i) - \epsilon_{\phi}(x_{t_i}; \emptyset, t_i))\frac{\partial g(\theta,c)}{\partial \theta}\right]$
        \State Update $\theta$
        \If{$iter\%iter_f=0$}
        \State $Score_k = \sum_{j=1}^{H\times W \times M} \frac{V(k)}{D(r_j,k)^2\times maxV(r_j)} $
        \State $Sort(Score_k)$
        \State Delete last $\eta$ 3D Gaussians
        \EndIf
    \EndFor
    \State Generate $K$ images $\hat{x}_{0}^t$ using $\hat{x}_{0}^t = \frac{x_{t} - \sqrt{1 - \bar{\alpha}_t} \epsilon_{\phi}(x_{t}; y, t)}{\sqrt{\bar{\alpha}_t}}$ by sampling timestep $t \in (0,200)$ from different camera poses.
    \State Generate detailed and plausible textures by $L_{rec} = \sum_i ||g(\theta , c_i) - \hat{x}_{i0}^t||_2$.
    \If{$n$ is $S$}
        \State Save 3D Gaussian Representation of the entire scene
        \State break
    \EndIf
    \State Save 3D Gaussian Representation $obj_n$ of text $y_n$
    \State $world(x) = r_n\cdot s_n \cdot obj_n(x) + t_n$ 
    \State Add $obj_n$ to the Scene by coordinate $world(x)$
\EndFor
\end{algorithmic}
\end{algorithm}

\newlength{\myrowheight}
\setlength{\myrowheight}{7.5\baselineskip}  

\begin{table*}[t]
\centering
\caption{Comparison of training and evaluation camera pose strategies across different methods.}
\renewcommand{\arraystretch}{1.2}
\begin{tabular*}{\textwidth}{@{\extracolsep{\fill}} l p{7.2cm} >{\centering\arraybackslash}p{6.5cm} }
\toprule
\textbf{Method} & \textbf{Training Camera Poses} & \textbf{Evaluation Camera Poses} \\
\midrule
\textbf{Text2Room}~\cite{hollein2023text2room} & 
Training camera poses are sampled along a predefined continuous trajectory. Camera orientations are adjusted based on heuristic tilt and rotation rules defined in the original implementation, allowing moderate variation in viewpoint along the path.
& 
\multirow{5}{*}{
  \centering
  \rule{0pt}{\myrowheight}
  \parbox[c][\myrowheight][c]{6.2cm}{
    the camera starts from the scene center, first moves along straight paths in multiple directions across the scene, and then performs a circular motion around the center, with the radius of the circular path set to two-thirds of the scene diameter.
  }
} \\

\textbf{Text2NeRF}~\cite{zhang2024text2nerf} & 
The cameras are placed within the scene, facing outward, and sampled within a spherical region with a ±60° pitch angle.

 & \\

\textbf{ProlificDreamer}~\cite{wang2024prolificdreamer} & 
Same as Text2NeRF; uses object-centric sampling without scene layout awareness. & \\

\textbf{Set-the-Scene}~\cite{cohen2023set} & 
The entire scene is placed at the center, and camera poses are sampled from the bounding sphere, with the cameras oriented toward the interior of the scene. & \\

\textbf{DreamScene (Ours)} & 
(1) Sample camera poses near the scene center within a constrained radius;  
(2) For indoor scenes, divide space into regions based on object layout and sample camera poses within each region;  
for outdoor scenes, sample camera poses along concentric circles with fixed angular direction;  
(3) Combine all sampled poses across stages.  & \\
\bottomrule
\end{tabular*}
\label{tab:camera_pose_comparison}
\end{table*}

\begin{figure}
\centering
\subfloat[\footnotesize SDS~\cite{poole2022dreamfusion}]{\includegraphics[width=0.20\textwidth]{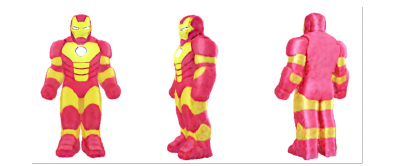}}
\subfloat[\footnotesize DreamTime~\cite{huang2023dreamtime}]{\includegraphics[width=0.20\textwidth]{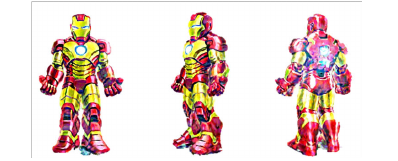}} \\
\subfloat[\footnotesize MTS]{\includegraphics[width=0.20\textwidth]{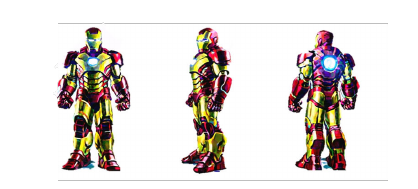}}
\subfloat[\footnotesize FPS]{\includegraphics[width=0.20\textwidth]{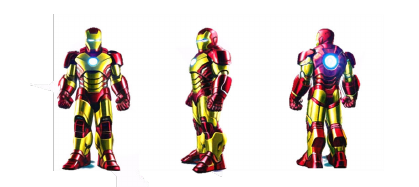}}

\caption{The ablation results of different sampling strategies. }
\label{fig:gen_by_recon}
\end{figure}

\section{Additional Experiments}
\subsection{Camera Configuration in Training and Testing}
To ensure fair and meaningful comparison across methods, we analyze the training-time camera pose strategies of existing baselines and apply a unified testing-time trajectory for all. Tab.~\ref{tab:camera_pose_comparison} provides a detailed comparison of training and evaluation camera pose strategies.

As shown in Table, for training-time camera poses, each baseline employs a distinct sampling strategy based on its architectural design:
\begin{itemize}
\item \textbf{Text2Room}~\cite{hollein2023text2room} samples camera poses along a predefined continuous trajectory. Camera orientations are adjusted using heuristic tilt and rotation rules from the original implementation, allowing moderate viewpoint variation along the path.

\item \textbf{Text2NeRF}\cite{zhang2024text2nerf} and \textbf{ProlificDreamer}\cite{wang2024prolificdreamer} place cameras within the scene, facing outward, and sample them within a spherical region constrained by a ±60° pitch angle.

\item \textbf{Set-the-Scene}~\cite{cohen2023set} centers the scene within a bounding sphere and samples camera poses from its surface, orienting the cameras inward toward the scene’s interior.
\end{itemize}

To enable a more fair and meaningful comparison, we adopt a unified camera trajectory during evaluation for all methods. Specifically, we test on the same scenes used for training but replace each method’s original training-time poses with a continuous camera trajectory that mimics natural human exploration behavior. The camera begins at the scene center, moves along straight paths in multiple directions across the environment, and then performs a circular sweep around the center. The radius of this circular path is set to two-thirds of the scene diameter. This unified trajectory better reflects real-world usage patterns and offers a more reliable measure of robustness and practical usability.

\subsection{Multi-head Scene}
\label{sec:multi-head}
\textcolor{black}{Fig.8 in the main paper illustrates the "multi-head" phenomenon observed in ProlificDreamer~\cite{wang2024prolificdreamer} and Text2Room~\cite{hollein2023text2room}. For methods relying on SDS~\cite{poole2022dreamfusion,lin2023magic3d,wang2024prolificdreamer}, the camera pose is randomly sampled during the optimization process, and the model lacks the ability to perceive the orientation of the scene. Consequently, the same prompt is optimized in any direction, often leading to the repetitive generation of objects, such as sofas, from all angles in scenarios like ”a living room“, resulting in an overwhelming presence of sofas in the final scene. For inpainting-based methods~\cite{zhang2024text2nerf,hollein2023text2room,ouyang2023text2immersion}, the model retains some orientation awareness as it continuously expands on a fixed-size rendering image—rotating a certain degree each time and completing it with the diffusion model. In cases where sofas have already appeared, these methods usually do not generate the same content again. However, if the rotation angle is large enough that the sofa disappears from the original view, the method will regenerate the sofa content. Overall, the "multi-head" issue is more pronounced with methods based on SDS than with inpainting-based methods. In our approach, because the scene layout is predefined, our model can utilize the orientation information and the existing object layout to enhance the environmental generation. This significantly mitigates the "multi-head" problem by ensuring that the environment generation is coherent and contextually appropriate.}
\subsection{Ablations}
\par \noindent  \textbf{Different sampling strategies.} 
We examined the effects of different sampling strategies on the generation results of a 3D object. Fig.~\ref{fig:gen_by_recon} (c) displays the outcomes after 30 minutes of optimization under the prompt "A DSLR photo of Iron Man." As shown, multi-timestep sampling (MTS) establishes superior geometric structures and textures compared to both the monotonically non-increasing sampling strategy in \cite{huang2023dreamtime}  and Score Distillation Sampling (SDS) technique in \cite{poole2022dreamfusion}. Building upon the strengths of MTS, Formation Pattern Sampling (FPS) employs a reconstruction method to produce smoother and more realistic textures.
\par
\begin{figure}
    \centering
    \includegraphics[width=\linewidth]{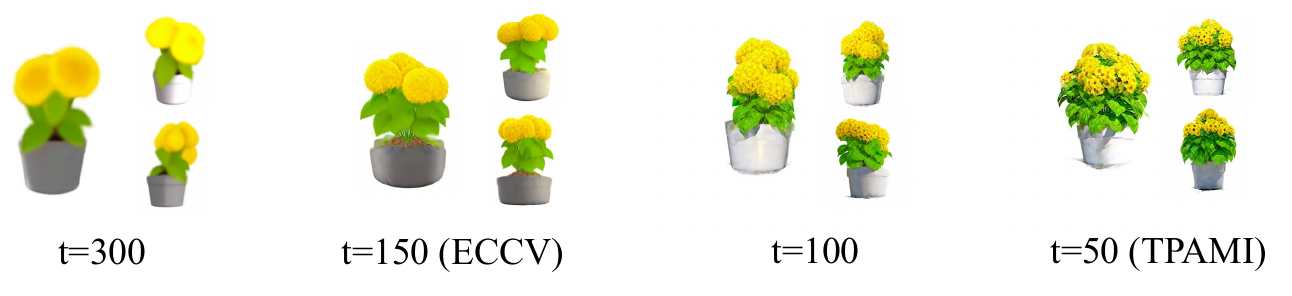}
    \caption{The ablation results of different timestep size $\Delta T$. }
    \label{fig:stepsize}
\end{figure}

 \noindent \textbf{Different sampling step sizes}. We conduct ablation studies using different sampling step sizes $\Delta T = t_{i+1} - t_i$. As shown in Eq.\ref{eq:approximation} and discussed in Sec.\ref{sec:Approximation}, smaller values of $\delta T$ lead to higher-quality and more detailed generation results, as further evidenced in Fig.~\ref{fig:stepsize}. However, smaller step sizes require more sampling steps. Considering hardware limitations, we adopt $\delta T$ values in the range of $50$ to $100$ in our experiments.

\end{document}